\documentclass[11pt]{article}

\usepackage[final]{acl}

\usepackage{times}
\usepackage{latexsym}
\usepackage{booktabs}
\usepackage{longtable}
\usepackage{multirow}
\usepackage{longtable}
\usepackage{array}
\usepackage{graphicx}
\usepackage{enumitem}
\usepackage[most]{tcolorbox}
\usepackage{xcolor}
\usepackage{tabularx}
\usepackage{array}
\usepackage{float}
\usepackage{amsmath}
\usepackage{amssymb}
\usepackage{caption}

\usepackage{afterpage}

\usepackage[T1]{fontenc}

\usepackage[utf8]{inputenc}

\usepackage{microtype}

\usepackage{inconsolata}

\usepackage{graphicx}

\usepackage[most]{tcolorbox}
\usepackage{caption}

\newtcolorbox{promptbox}[1]{
    enhanced,
    breakable,
    colback=white,
    colframe=black!55,
    colbacktitle=black!8,
    coltitle=black,
    fonttitle=\bfseries,
    title={#1},
    boxrule=0.5pt,
    arc=1mm,
    left=6pt,
    right=6pt,
    top=6pt,
    bottom=6pt,
    before skip=8pt,
    after skip=8pt}

\newcommand{\corpusOne}{134{,}213} 
\newcommand{\corpusTwo}{133{,}813} 
\newcommand{\corpusmisinfo}{1{,}730} 

\title{Beyond Binary Detection: A Multi-Dimensional Taxonomy of Cancer Misinformation on Reddit}

\author{
 \textbf{Aria Pessianzadeh\textsuperscript{1}},
 \textbf{Pooriya Jamie\textsuperscript{2}},
 \textbf{Naima Sultana\textsuperscript{1}},
 \textbf{Georgia Himmelstein\textsuperscript{3}},
\\
 \textbf{Yuliya Zektser\textsuperscript{3}},
 \textbf{Patricia Ganz\textsuperscript{3}},
 \textbf{Homa Hosseinmardi\textsuperscript{2}},
 \textbf{Amir Ghasemian \textsuperscript{2}},
\\
 \textbf{Rezvaneh Rezapour\textsuperscript{1}}
\\
 \textsuperscript{1}Department of Information Science, Drexel University\\
 \textsuperscript{2}OASIS Lab, UCLA\\
 \textsuperscript{3}Division of Hematology and Oncology, Department of Medicine, UCLA Health
\\{
   \texttt{ap3943@drexel.edu, shadi.rezapour@drexel.edu}
 }
}

\begin{document}
\maketitle
\begin{abstract}
Cancer-related discussions on social media provide important spaces for information exchange and peer support, but can also expose users to misinformation with implications for prevention, screening, and treatment decisions. Existing work often treats cancer misinformation as a binary phenomenon, providing limited insight into how misinformation is expressed, engaged with, and associated with potential harm. We introduce a multi-dimensional taxonomy for characterizing cancer misinformation in Reddit discussions of breast, lung, colon, and prostate cancer. Developed through expert annotation, the taxonomy captures seven dimensions spanning misinformation presence, cancer stage, information-seeking and sharing behavior, misinformation type, risk, stance, and topical focus. We evaluate 21 large language models (LLMs) across zero- and few-shot settings and develop a cross-model agreement strategy for scaling misinformation identification to over 133K posts. Our analysis shows that misinformation is heterogeneous in both form and function: unproven and alternative treatments emerge as a prominent topic, misinformation frequently occurs within exchanges that combine information seeking and sharing, and users engage with questionable claims through both endorsement and uncertainty. We further find that classification difficulty varies substantially across dimensions, with risk assessment posing particular challenges for both human annotators and LLMs. Our taxonomy and empirical findings move beyond binary detection toward a more nuanced characterization of how cancer misinformation is produced, discussed, and encountered in online health communities.

\end{abstract}

\section{Introduction}\label{sec:intro}

The rapid growth of social media platforms has transformed how individuals seek, share, and engage with health-related information~\cite{moorhead2013new,jia2021online,bouzoubaa2023exploring}. Beyond serving as sources of medical information, these platforms also provide spaces for emotional support \cite{pessianzadeh2026reddit}, experiential knowledge sharing \cite{phadke2026reinforcing}, and discussions about health conditions and treatment options \cite{aghakhani-rezapour-2026-like}. These functions are particularly important in cancer communities, where individuals often navigate substantial uncertainty, emotional vulnerability, and complex care decisions~\cite{harkin2020secret, tripathi2022examination}. At the same time, social media facilitates the circulation of cancer-related misinformation~\cite{johnson2022cancer,loeb2024cancer,ghenai2018fake}, which may be amplified by these same contextual factors~\cite{swire2024cancer,wu2022linking,wang2019systematic}. Exposure to misleading or inaccurate information about cancer prevention, screening, diagnosis, or treatment can delay care, reduce adherence to evidence-based interventions, and increase the adoption of harmful or ineffective therapies~\cite{loeb2024cancer,johnson2022cancer}. Moreover, cancer-related misinformation often receives disproportionately high levels of online engagement~\cite{loeb2024cancer,meteran2023usefulness,johnson2022cancer}.

Prior work has documented cancer misinformation involving unproven treatments~\cite{fridman2025identifying}, 
dietary interventions~\cite{warner2022online}, 
supplements~\cite{wilner2020breast}, miracle cures, and conspiratorial narratives targeting pharmaceutical companies and medical 
institutions~\cite{warner2021young, varet2025assessing}. Other studies have examined how such content is communicated through emotional narratives \cite{trivedi2022factors}, anecdotal evidence \cite{loeb2024cancer}, and distrust framing~\cite{peng2023persuasive}. Parallel research in NLP and computational social science has examined cancer-related social media discussions through the lens of stigma, emotional expression, anxiety, and social support~\cite{rajwal2024unveiling, lal2024analysing, pierce2024identifying, liu2024breaking}. 

Despite these advances, existing work on cancer misinformation still lacks a unified computational framework for large-scale characterization. Prior studies often rely on qualitative analyses, small-scale datasets, or binary misinformation labels that fail to capture how misinformation is framed, contextualized, or engaged with~\cite{swire2024cancer}. As a result, important distinctions between fabricated claims, misleading interpretations, anecdotal evidence, and contextualized uncertainty remain underexplored~\cite{wardle2017information}.
To address these limitations, we develop a multi-dimensional taxonomy for cancer misinformation grounded in established misinformation typologies and refined through expert feedback~\cite{wardle2017information}. The framework captures seven dimensions: Misinformation Presence, Cancer Stage, Information Type, Misinformation Type, Risk Level, Stance on Misinformation, and Misinformation Topic. Using this taxonomy, we construct an expert-annotated dataset and systematically evaluate 21 LLMs across zero-shot and few-shot settings. We then develop a cross-model agreement strategy for scalable misinformation identification and apply the framework to more than 133K Reddit posts spanning breast, lung, colon, and 
prostate cancer communities.

Our findings demonstrate both the promise and challenges of fine-grained LLM-based cancer misinformation classification. The selected classifiers achieved weighted F1 scores of approximately 0.89 and high sensitivity (TPR = 0.98). Their intersection identified 1,730 high-confidence misinformation posts (1.29\% of the analysis corpus) for downstream characterization. Performance varied across taxonomy dimensions, with Misinformation Topic and Cancer Stage achieving stronger performance, while more subjective dimensions, particularly Risk Level, remained challenging for both human annotators and LLMs.
Our analysis further shows that cancer misinformation varies substantially in form and communicative function. Unproven or alternative treatments emerged as the most prevalent topic, while nearly half of identified posts combined information-seeking and sharing, and stance patterns reflected both endorsement and uncertainty. We also observed variation across cancer communities and recurring evidence themes involving unsupported treatments, misconceptions about cancer causes and processes, and distrust of conventional medicine.

Our contributions are threefold. First, we introduce a comprehensive multi-dimensional taxonomy that moves beyond binary detection to characterize the form, context, potential risk, stance, and topical focus of cancer misinformation. Second, we develop and release an expert-informed annotated dataset and systematically evaluate LLM-based approaches for scaling these annotations, including a cross-model agreement strategy for identifying high-confidence misinformation. Third, we provide a large-scale computational analysis of cancer misinformation across Reddit cancer communities, revealing how misinformation varies in its content, communicative function, and user engagement. Our work provides a foundation for fine-grained NLP analysis of health misinformation in online communities. Data, code, and 
results can be found here: \url{https://github.com/social-nlp-lab/cancer-misinformation}.

\section{Related Work} \label{sec:relatedwork}

\noindent\textbf{Conceptualizing and Characterizing Cancer Misinformation. }
Recent work highlights cancer as a critical domain for misinformation research due to its high emotional stakes, treatment uncertainty, and serious behavioral consequences~\cite{swire2024cancer}. Within this literature, cancer misinformation is generally defined as health-related content that is false, unsupported, misleading, or inconsistent with current scientific consensus~\cite{lazard2023exposure, johnson2022cancer}. Prior studies have identified misinformation in unproven treatments, unsupported prevention strategies, and exaggerated intervention claims~\cite{loeb2024cancer,yussof2023breast,fridman2025identifying}, including alternative therapies, supplements, herbal cancer cures, and food and lifestyle claims~\cite{yussof2023breast,fridman2025identifying}. Existing work further suggests that cancer misinformation is not monolithic. \citet{yussof2023breast} distinguished between inaccurate and fabricated misinformation, while \citet{fridman2025identifying} showed that misleading content often combined factual and false claims, complicating binary true/false annotation. This distinction is further complicated by debunking, sarcasm, or neutral discussion of alternative medicine, where mention does not imply endorsement~\cite{fridman2025identifying}.

Beyond topical categories, cancer misinformation is often embedded in broader narrative and persuasive forms. Common patterns include miracle-cure advice, discouragement of conventional treatment, and distrust-centered narratives targeting doctors or pharmaceutical companies~\cite{warner2021young}. Such claims may also employ persuasive strategies such as personal narratives, institutional distrust, and metaphoric framings (e.g., detoxification or combat)~\cite{peng2023persuasive,kamali2024using,loeb2024cancer,lazard2026intervening,baker2025conspiracy}. These findings suggest that cancer misinformation should be understood not only as inaccurate content, but also as discourse whose meaning and potential impact depend on how claims are framed and communicated~\cite{komsany2026leveraging}.

\noindent\textbf{Misinformation Harms and Gaps.}
Cancer misinformation is consequential not only because it is inaccurate, but because it can produce tangible harm. Identified harms include delaying medical care, adopting ineffective or toxic treatments, economic harm, and harmful interactions with legitimate therapies~\cite{loeb2024cancer}. Accordingly, harmful content has been evaluated not only by factual accuracy, but also by factors such as actionability, believability, spread potential, and exploitative capacity~\cite{sehat2024misinformation,tran2019investigation,tran2020misinformation}. Related health misinformation work further highlights the role of rumors, incomplete information, and skepticism toward medical guidance in shaping harmful interpretations and behaviors~\cite{alvarez2025development}.

At the same time, NLP research on online cancer discussions has focused on adjacent problems rather than misinformation itself~\cite{podina-etal-2021-natural, podina2023mental}. Existing work examined patient concerns, emotional expression, entity extraction, anxiety during diagnosis, stigma in lung cancer discussions, and cancer-survivor discourse on Reddit~\cite{rajwal2024unveiling,lal2024analysing,lal2025lens,pierce2024identifying,liu2024breaking,podina-etal-2021-natural}. This work demonstrates the value of NLP for cancer-related social media analysis, but lacks a unified, multi-dimensional characterization of misinformation at scale. We address this gap with a taxonomy and computational framework for large-scale analysis of cancer misinformation on Reddit. 
\section{Method} \label{sec:method}

\subsection{Data Collection}
Reddit provides a valuable environment for studying health discourse due to its long-form, community-driven discussions that capture patient experiences and information exchange~\cite{de2014mental,sarker2015utilizing}. To examine cancer misinformation, we collected discussions from communities associated with four cancer types selected in consultation with domain experts: \textit{breast}, \textit{lung}, \textit{colon}, and \textit{prostate}. We identified relevant communities, resulting in 24 subreddits. Data were retrieved using Arcticshift~\cite{Heitmann2025arctic_shift}, an interface to the historical Pushshift archives~\cite{Baumgartner2020Pushshift}. In total, we collected 243,538 posts. We removed empty posts, posts with fewer than five words, content authored by automated accounts (e.g., ``AutoModerator''), and posts marked as deleted or removed to protect user privacy, resulting in a final dataset of \corpusOne{} posts (see Table~\ref{tab:subreddits list} for the full list of subreddits). After excluding the 400 posts used for human annotation, \corpusTwo{} posts remained for full-corpus inference and downstream analysis.

\subsection{Taxonomy of Cancer Misinformation}
We constructed a multi-dimensional framework in collaboration with domain experts to analyze cancer-related misinformation on Reddit. Rather than treating misinformation as a binary phenomenon, we characterize its structure, communicative role, and potential impact. We developed an annotation taxonomy consisting of seven complementary dimensions (Table \ref{tab:taxonomy}). 

\noindent\textbf{Misinformation Presence.}
This dimension captures whether a post contains cancer-related misinformation. Following prior health misinformation research~\cite{chou2018addressing}, we define cancer misinformation as false, misleading, or scientifically unsupported claims about cancer risk factors, prevention, diagnosis, or treatment that may lead to harmful consequences.
We additionally capture the specific textual \textit{evidence} supporting the misinformation classification.

\noindent\textbf{Information Type.}
Posts are categorized as information-seeking, information-sharing, or both, reflecting different forms of health communication behavior~\cite{lambert2007health}. Seeking includes requests for advice, clarification, or second opinions, while sharing includes providing information, experiences, opinions, or claims.

\begin{table}[ht]
\centering
\small
\resizebox{\columnwidth}{!}{
\begin{tabular}{p{3.5cm} p{5.5cm}}
\toprule
\textbf{Dimension} & \textbf{Categories} \\
\midrule

Misinformation Presence
& Yes; No; Unclear \\

Cancer Stage
& Prevention / Screening; Treatment; Other/ Unclear / Not applicable \\

Information Type 
& Seeking; Sharing; Both; Neither / Unclear \\

Misinformation Type 
& Satire or Parody; False Connection; Misleading Content; False Context; Imposter Content; Fabricated Content; Manipulated Content; Other \\

Risk Level 
& Low Risk; Moderate Risk; High Risk \\

Stance on Misinformation 
& Accept; Accept + Question; Reject; Reject + Question; Question; Other/Unclear \\

Misinformation Topic 
& Causation, risk factors \& prevention claims; Diagnosis \& screening claims; Safety \& effectiveness of conventional treatments; Unproven or alternative treatments; Conspiracy \& institutional distrust; Morality, religion \& ideology; Medical autonomy \& medical freedom; Other \\

\bottomrule
\end{tabular}
}
\caption{Overview of the multi-dimensional taxonomy for characterizing cancer misinformation.}
\label{tab:taxonomy}
\vspace{-0.5cm}
\end{table}

\noindent\textbf{Cancer Stage.}
This category distinguishes discussions related to cancer screening, treatment, or neither. Prior work shows that misinformation can influence both preventive behaviors and treatment adherence, often with different downstream consequences~\cite{nagler2014adverse,chou2018addressing}.

\noindent\textbf{Misinformation Type.}
We distinguish between seven forms of misinformation: satire/parody, false connection, misleading content, false context, imposter content, fabricated content, and manipulated content. Posts that do not clearly fit these types are assigned to an additional \textit{Other} category. This taxonomy is adapted from established misinformation typologies that differentiate mechanisms of deception and distortion~\cite{wardle2017fakenews,wardle2017information}.

\noindent\textbf{Risk Level of Misinformation.}
We assess misinformation severity based on its potential impact on health behavior, ranging from low to high risk. Prior research shows that health misinformation varies substantially in its consequences, from relatively benign misconceptions to claims that may discourage effective treatment or promote harmful interventions~\cite{southwell2019misinformation, wang2019systematic, van2022misinformation}. Incorporating risk enables distinguishing misinformation by its potential for harm.

\noindent\textbf{Stance on Misinformation.} 
This category captures how users engage with misinformation-related content. Following prior work on stance detection~\cite{kuccuk2020stance,rezapour2021incorporating}, we distinguish between accepting, rejecting, and questioning misinformation. Accepting includes supporting, promoting, or endorsing misleading claims, while rejecting reflects critique, correction, or explicit disagreement. Questioning captures uncertainty, requests for clarification, or exploratory discussion around potentially misleading content. Building on~\cite{pessianzadeh2025exploring}, we additionally include hybrid categories where acceptance or rejection co-occurs with questioning to capture mixed or ambiguous positions.

\noindent\textbf{Misinformation Topic.}
This category captures the thematic focus of cancer-related misinformation. Drawing on prior work in health misinformation~\cite{kata2010postmodern}, we distinguish between eight topical categories. These categories enable analysis of how narratives vary across domains such as treatments, prevention strategies, and skepticism toward conventional medicine.

\subsection{Data Annotation}
Since cancer misinformation is relatively sparse in online discussions, fully random sampling would produce a dataset dominated by non-misinformation posts and inefficiently use limited domain-expert annotation resources. Following prior findings on the low prevalence of health misinformation~\cite{suarez2021prevalence, broniatowski2020twitter}, we adopted a stratified sampling strategy to construct a more informative annotation set. We first used Qwen, an open-weight LLM, to weakly label posts for the presence or absence of misinformation. We then applied GPT-5.4, GPT-5.2, and Claude-Haiku-4.5 to Qwen-positive posts and an equally sized random sample of Qwen-negative posts to identify high-confidence candidate pools based on cross-model agreement. From the resulting consensus pools, we sampled 200 consensus-positive and 200 consensus-negative posts for human annotation, resulting in a stratified annotation set of 400 posts. 

\begin{figure}[t]
    \centering
    \includegraphics[width=0.75\linewidth]{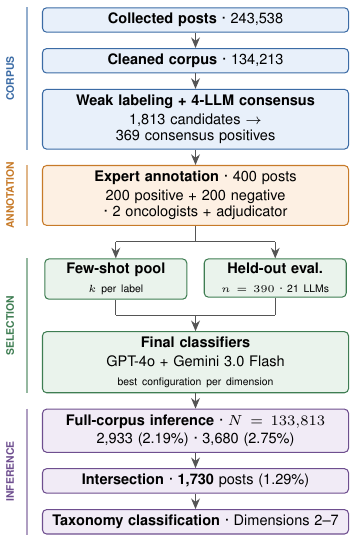}
    \caption{Overview of the methodology (data selection, annotation, and classification).}
    \label{fig:data_annotation}
\vspace{-0.55cm}
\end{figure}

The annotation process was conducted by two domain experts in cancer care. We used a hierarchical workflow in which annotators first identified whether a post contained misinformation; only posts labeled as misinformation were further annotated across the remaining taxonomy dimensions. Disagreements between the two primary annotators were adjudicated by a third experienced oncologist to obtain the final labels. Annotation was conducted using Potato~\cite{pei2022potato}, with a detailed codebook defining each taxonomy category. Figure~\ref{fig:data_annotation} summarizes the data selection, annotation, and classification pipeline; further details are provided in Appendix~\S\ref{sec:weak_labeling}.

\subsection{Classification}
Following the proposed annotation framework, we evaluated LLMs for cancer misinformation classification using the same hierarchical structure as the annotation process. 
Models first identified posts containing misinformation, which were then classified across the remaining taxonomy dimensions.

\subsubsection{Misinformation Classification} 
\noindent\textbf{Models.} We evaluated 21 LLMs on our ground-truth data. We selected models from OpenAI \cite{brown2020language}, Claude 
\cite{anthropic2026claude}, Gemini 
\cite{team2023gemini}, Meta-Llama-3.1-8B \cite{grattafiori2024llama}, Gemma-4-31B-it \cite{team2024gemma}, and Qwen3.6-27B \cite{yang2025qwen3} to compare model families (see Table \ref{tab:model_performance} for full list of models).

\noindent\textbf{Prompting.} We evaluated models in zero-shot and few-shot settings. In the zero-shot setting, models received the task instructions and our definition of misinformation and classified each post as either misinformation or other, with the latter encompassing non-misinformation and unclear cases. For few-shot prompting, we used $k \in [1,5]$ human-annotated instances per class, drawn from a fixed pool of 10 posts excluded from all evaluations, resulting in a held-out set of $N=390$. In both settings, we incorporated Chain-of-Thought (CoT) prompting~\cite{wei2022chain}, asking models to provide a rationale for their predictions and, when misinformation was identified, extract the supporting textual evidence. We used this evidence for the subsequent analysis in \S\ref{sec:evidence_analysis}. Complete prompts are provided in \S\ref{sec:taxonomy_prompts}.

\subsubsection{Taxonomy Dimension Classification}
For posts identified as misinformation, we further evaluated LLM performance across the remaining taxonomy dimensions: Cancer Stage, Information Type, Misinformation Type, Risk Level, Stance, and Misinformation Topic. We evaluated models in zero- and few-shot settings. For few-shot prompting, we used $k \in [1,4]$ human-annotated instances per class, selected from our ground-truth data and excluded from the corresponding evaluation set. Prompts were constructed separately for each taxonomy dimension and shot setting.

\subsubsection{Experiment Setting}
Models were evaluated using weighted precision (P), weighted recall (R), and weighted $F_1$. For binary misinformation classification, we report the true positive rate (TPR) for misinformation and the true negative rate (TNR) for non-misinformation. Across all experiments, model temperature was set to 0 to reduce output variability. Unless otherwise specified, all remaining generation parameters were kept at their default values for each model API. 

\section{Results} \label{sec:result}
\begin{table}[t]
\centering
\small

\begin{tabular}{lcc}
\toprule
\textbf{Dimension} & \textbf{Cohen's $\kappa$} & \textbf{AC1} \\
\midrule
Misinformation Presence   & 0.85 & 0.90 \\ 
Cancer Stage             & 0.48 & 0.70 \\
Information Type                & 0.68 & 0.79 \\
Misinformation Type          & 0.30 & 0.60 \\
Risk Level                      & 0.30 & 0.32 \\
Stance on Misinformation    & 0.49 & 0.59 \\
Misinformation Topic	        & 0.48 & 0.63 \\
\bottomrule
\end{tabular}
\caption{Inter-annotator agreement between the two primary annotators across taxonomy dimensions, measured using Cohen's $\kappa$ and Gwet's AC1.}
\label{tab:kappa_scores}
\vspace{-0.5cm}
\end{table}
\subsection{Annotation Agreement} \label{sec:annotation_agreement}
Two domain experts independently annotated 400 sampled posts. 
Table~\ref{tab:kappa_scores} reports inter-annotator agreement across the seven annotation dimensions using 
Cohen's $\kappa$ and Gwet's AC1. We report AC1 alongside $\kappa$ because it is less sensitive to skewed marginal distributions and class imbalance, which are present in several of our multi-class dimensions.
Agreement was highest for Misinformation Presence ($\kappa = 0.85$, AC1 = 0.90), followed by Information Type ($\kappa = 0.68$, AC1 = 0.79). AC1 was notably higher than $\kappa$ for several dimensions, particularly Cancer Stage and Misinformation Type, suggesting that lower $\kappa$ values partly reflect uneven label distributions rather than disagreement alone. Risk Level showed the lowest agreement under both measures ($\kappa = 0.30$, AC1 = 0.32); unlike the other lower-$\kappa$ dimensions, agreement did not increase substantially under AC1, suggesting greater difficulty in consistently assessing the potential severity and real-world consequences of misinformation. 
Following the agreement analysis, disagreements between the two primary annotators were adjudicated by a third experienced oncologist to obtain a single final label for each post. The third annotator's labels were used only for adjudication and excluded from agreement statistics. Tables~\ref{tab:misinformation_examples} and~\ref{tab:task_examples} present representative examples across dimensions and labels. Additional details and examples of annotator disagreement are provided in ~\S\ref{sec:weak_label_validation} and \S\ref{sec:annotator_disagreement}.

\noindent\textbf{Ground-Truth Label Distribution.} Following adjudication, the ground-truth dataset contained 161 misinformation (40.25\%), 235 non-misinformation (58.75\%), and 4 unclear posts (1.00\%). This distribution reflects our consensus-enriched, stratified sampling procedure and should not be interpreted as corpus-level prevalence. The remaining dimensions were annotated only for the 161 misinformation posts. Treatment (70\%), Fabricated Content (72\%), Low Risk (58\%), and Unproven or Alternative Treatments (62\%) were the most frequent labels in their respective dimensions, while 52\% of posts combined information seeking and sharing. Full distributions are reported in Table~\ref{tab:annotation_distribution}.

\subsection{Misinformation Classification}
\begin{figure}[t]
    \centering
    \includegraphics[width=\linewidth]{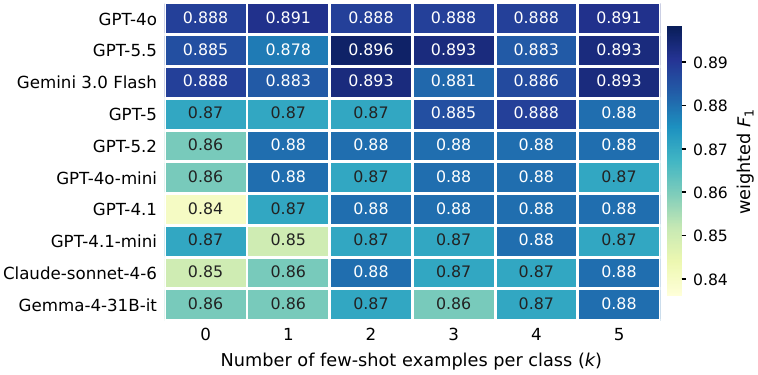}
    \caption{F1 scores for misinformation classification across LLMs for $k=0-5$.} 
    \label{tab:F1_Shot_Settings}
    \vspace{-0.5cm}
\end{figure}

We evaluated 21 LLMs across zero-shot and few-shot settings ($k=0$--$5$). Figure~\ref{tab:F1_Shot_Settings} shows F1 scores for selected models, with full results reported in Table~\ref{tab:model_performance}. Several models performed comparably well, with the strongest models remaining relatively stable across shot settings. In the zero-shot setting, GPT-5.5, GPT-4o, and Gemini 3.0 Flash achieved weighted F1 scores of 0.885, 0.888, and 0.888, respectively; however, GPT-4o and Gemini 3.0 Flash showed higher sensitivity to misinformation (TPR = 0.97) than GPT-5.5 (TPR = 0.91). Across few-shot settings, the highest weighted F1 was 0.896, achieved by GPT-5.5 at $k=2$, followed closely by Gemini 3.0 Flash (F1 = 0.893 at $k=2,5$) and GPT-4o (F1 = 0.891 at $k=1,5$).

Despite its slightly higher F1, GPT-5.5's higher inference cost and generally lower sensitivity made it less suitable for our corpus-scale setting. Based on classification performance, sensitivity to misinformation, stability across shots, and inference cost, we selected GPT-4o ($k=1$; F1 = 0.891, TPR = 0.98) and Gemini 3.0 Flash ($k=5$; F1 = 0.893, TPR = 0.98) as the final classifiers. Full model and cross-model evaluation details are provided in Appendix~\S\ref{sec:full_classificaion}.

\noindent\textbf{Overall Distribution of Misinformation}
The two selected classifiers were independently applied to the full analysis corpus ($N=\corpusTwo$). GPT-4o classified 2,933 posts (2.19\%) as misinformation, while Gemini 3.0 Flash classified 3,680 posts (2.75\%). Since the two models showed comparable performance on the human-annotated evaluation set, we used their intersection (i.e., posts independently classified as misinformation by both models) to construct the high-confidence misinformation set. Both models classified 1,730 misinformation posts (1.29\% of the corpus), which were retained for downstream analyses. The models disagreed on 3,153 posts (2.36\%), which were excluded from the high-confidence misinformation set. The intersection provides a conservative operationalization of misinformation rather than an estimate of its true prevalence. Detailed evaluation of the intersection and union decision rules, cross-model disagreement, and label-selection rationale are provided in \S\ref{sec:full_corpus_agreement}-\ref{sec:downstream_label_selection}.

\subsection{Taxonomy Dimension Classification}
We next evaluated GPT-4o and Gemini 3.0 Flash on the remaining taxonomy dimensions across zero-shot and few-shot settings ($k=0$--$4$). Table~\ref{tab:model_task_shot_performance} reports performance for each model. Results varied substantially across dimensions, reflecting differences in classification difficulty. Misinformation Topic had the strongest performance, with GPT-4o achieving a weighted F1 of 0.96 at $k=3$ and maintaining F1 scores of 0.94–0.96 across few-shot settings. Cancer Stage also achieved strong performance, with Gemini 3.0 Flash outperforming GPT-4o across settings and reaching an F1 of 0.90 at $k=3$. For Information Type, GPT-4o achieved the highest F1 of 0.87 at $k=1$. Few-shot prompting produced the largest gains for Misinformation Type and Risk Level. For Misinformation Type, GPT-4o improved from an F1 of 0.12 in the zero-shot setting to 0.71 at $k=4$. Similarly, performance on Risk Level also improved with few-shot prompting but remained challenging; the highest F1 was 0.40, achieved by Gemini 3.0 Flash at $k=3$. This lower performance is consistent with the lower human inter-annotator agreement observed for Risk Level (Table~\ref{tab:kappa_scores}), indicating that distinctions among risk categories were challenging for both human annotators and LLMs. For Stance on Misinformation, Gemini 3.0 Flash performed better overall, reaching a maximum F1 of 0.67 at $k=2$.

For downstream analysis, we selected the model and prompting with the highest F1 for each taxonomy dimension. The selected configurations were then applied to the high-confidence misinformation set for subsequent analyses.

\noindent\textbf{Taxonomy Dimension Distribution}
After applying the selected model and prompting configuration for each taxonomy dimension to the high-confidence misinformation set, we examined the distribution of labels across dimensions. Table~\ref{tab:taxonomy_distribution} presents these distributions and shows several patterns in cancer misinformation discussions on Reddit. Within the Cancer Stage dimension, treatment-related misinformation was the largest category (47.80\%), followed by prevention/screening (19.31\%), while 32.89\% of posts were classified as other, unclear, or not applicable. For Information Type, nearly half of the posts (48.84\%) combined information-seeking and information-sharing behaviors, compared with 30.06\% that primarily shared information and 21.10\% that primarily sought information. This suggests that misinformation frequently occurs in interactive exchanges where users both introduce claims or experiences and seek feedback from others. For Misinformation Type, false context (29.60\%) and fabricated content (29.36\%) were the most common categories, followed by other forms (27.40\%). 
For Risk Level, more than half of the posts were classified as moderate risk (54.34\%), while 15.84\% were classified as high risk. 
The Stance distribution further indicates that users did not engage with misinformation solely through explicit endorsement. Questions were the most frequent stance (32.14\%), followed closely by acceptance (29.48\%) and posts combining acceptance with questioning (20.40\%). 
These patterns suggest that misinformation frequently appears within a mixture of endorsement and uncertainty rather than through unequivocal acceptance alone. Finally, unproven or alternative treatments were most prevalent (44.98\%), substantially exceeding diagnosis and screening claims (16.88\%) and claims concerning the safety and effectiveness of conventional treatments (15.50\%). Causation, risk factors, and prevention claims accounted for 7.06\%, while conspiracy and institutional distrust (4.57\%), medical autonomy and medical freedom (2.20\%), and morality, religion, and ideology (1.27\%) were less common. As the results show, treatment-related misinformation, particularly claims involving unproven or alternative approaches, constitutes a prominent component of cancer misinformation in the Reddit discussions.

\subsection{Evidence Analysis}\label{sec:evidence_analysis}
Our LLM-based annotation pipeline additionally extracted the textual evidence supporting each misinformation classification. We applied LLooM~\cite{lam2024conceptInduction} to identify recurring themes within this evidence beyond the predefined taxonomy categories. Table~\ref{tab:reasoning_themes} presents the themes, their inclusion criteria, and representative examples. The themes span alternative and natural treatments, misconceptions about cancer causes and processes, cure-related conspiracies, misuse of cancer statistics, skepticism toward conventional medicine, emotional or psychological explanations of cancer, and purported links between vaccines and cancer. They also capture more specific misconceptions about cancer detection, severity, spread, and recurrence. For example, the evidence includes claims that natural remedies can treat cancer, that conventional treatments can promote cancer spread, or that vaccines can trigger cancer. 

\subsection{Misinformation Across Cancer Communities}
Table~\ref{tab:subreddit_distribution} shows the proportion of high-confidence misinformation posts across cancer-related subreddits. Large communities such as \textit{r/breastcancer}, \textit{r/Prostatitis}, and \textit{r/ProstateCancer} contained the highest absolute numbers of misinformation posts, consistent with their overall discussion volume. Among high-volume communities, \textit{r/lungcancer} exhibited the highest misinformation ratio (2.18\%), followed by \textit{r/ProstateCancer} (1.99\%) and \textit{r/Prostatitis} (1.91\%). In contrast, \textit{r/colonoscopy} (0.45\%) and \textit{r/doihavebreastcancer} (0.78\%) showed comparatively lower ratios. Several smaller niche communities exhibited substantially higher ratios of posts classified as misinformation, including \textit{r/Prostate\_Cancer} (50.00\%), \textit{r/ProstateCancer\_AS} (45.95\%), and \textit{r/BreastCancerResearch} (20.75\%), although these findings should be interpreted cautiously due to limited sample sizes. These results show substantial cross-community variation, although ratios from small subreddits should not be interpreted as reliable estimates of community-level misinformation prevalence.

\begin{figure}
    \centering
    \includegraphics[width=\linewidth]{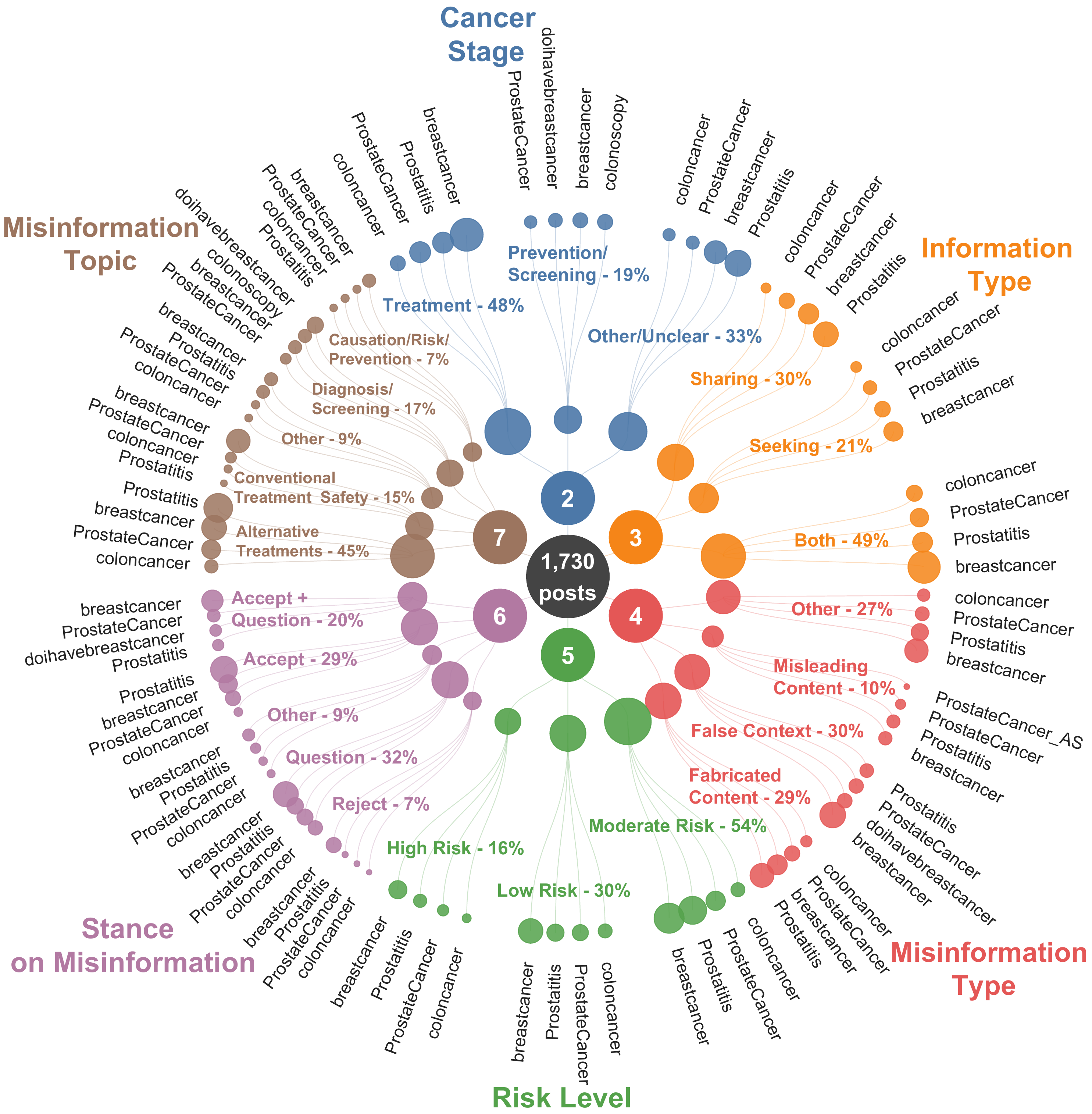}
    \caption{Breakdown of 1,730 high-confidence misinformation posts across six taxonomy dimensions, with the top four subreddits contributing to each displayed label shown in the outer ring. Labels representing less than 4.5\% of posts are omitted for readability, and Misinformation Topics are shortened for display.}
    \vspace{-0.5cm}
    \label{fig:radial}
\end{figure}

Figure~\ref{fig:radial} integrates these results, mapping the 1,730 high-confidence misinformation posts across six taxonomy dimensions and their contributing subreddits. Across many labels, the same high-volume communities, including \textit{r/breastcancer}, \textit{r/Prostatitis}, \textit{r/ProstateCancer}, \textit{r/coloncancer}, and \textit{r/doihavebreastcancer}, appear prominently. Treatment-stage misinformation, combined information seeking and sharing, moderate-risk content, questioning-oriented discourse, and unproven or alternative treatment claims appear across these active communities, showing that common taxonomy labels span multiple cancer communities and partly reflect subreddit size and activity.

\section{Discussion} \label{sec:discussion}

\noindent\textbf{Misinformation is Relatively Rare but Still Significant.}
Our conservative cross-model agreement strategy identified 1,730 high-confidence misinformation posts, representing 1.29\% of the analysis corpus. This proportion should not be interpreted as the true prevalence of cancer misinformation, as our intersection-based approach prioritizes classification confidence and may exclude cases identified by only one model. Prior work similarly shows that the prevalence of health misinformation varies substantially across social media contexts~\cite{suarez2021prevalence}. Nevertheless, the presence of more than 1,700 high-confidence instances illustrates that misinformation persists even within communities largely centered on legitimate information exchange and peer support. This is particularly important in cancer communities, where unsupported claims may shape consequential decisions about prevention, screening, and treatment. Such claims may be especially consequential when embedded within personal experiences, information-seeking exchanges, or peer-to-peer advice~\cite{peng2023persuasive,swire2024cancer}. These findings highlight the importance of examining how misinformation is communicated and encountered, even when relatively infrequent.

\noindent\textbf{Moderation Challenges in Uncertainty-Oriented Health Discussions.}
Although many cancer-related subreddits are relatively specialized and actively moderated, misinformation can persist through conversational and uncertainty-oriented discourse, including anecdotal experiences, unproven or alternative treatments, and misleading interpretations of medical evidence. Such content is difficult to moderate because it often blends personal experience, questions, and partial truths rather than presenting explicit falsehoods. Stance patterns reflected a mixture of questioning (32.14\%), acceptance (29.48\%), and acceptance combined with questioning (20.40\%). Nearly half of identified posts (48.84\%) also combined information seeking and sharing, indicating that users may introduce claims or experiences while simultaneously seeking feedback or validation from others. These patterns suggest that misinformation can emerge through interactive sensemaking and uncertainty as well as direct endorsement~\cite{wardle2017information,cinelli2020covid,vraga2020correction}.
In Reddit communities, where moderation is largely community-driven and discussion norms emphasize peer support and open conversation, overly aggressive moderation may suppress legitimate health concerns, while insufficient moderation may allow misleading narratives to normalize within support-oriented discussions. This tension motivates context-sensitive moderation that account for misinformation type, stance, risk level, and conversational intent rather than relying solely on binary true/false classification.

\noindent\textbf{Subjectivity of Some Taxonomy Dimensions.}
Several dimensions of our taxonomy exhibited greater annotation and modeling difficulty, with Risk Level showing the greatest difficulty: low agreement under both Cohen's $\kappa$ and Gwet's AC1 and the weakest downstream classification performance. For other dimensions, such as Misinformation Type, the higher AC1 relative to $\kappa$ suggests that uneven label distributions contributed to lower observed agreement. More broadly, some fine-grained dimensions require nuanced judgments about intent, ambiguity, rhetorical framing, and potential behavioral consequences. This reflects broader challenges in misinformation research, where disagreement can arise not only from factual uncertainty but also from differences in contextual interpretation and perceptions of harm~\cite{wang2019systematic,van2022misinformation}. Such ambiguity is further amplified by the coexistence of personal experience, emotional support seeking, evolving medical evidence, and uncertainty around treatment decisions. Our results therefore suggest that multi-dimensional misinformation analysis introduces annotation and modeling challenges beyond binary detection, particularly for dimensions requiring contextual or normative judgment.

\noindent\textbf{Frequency and Potential Harm are Distinct.}
More than half of the identified misinformation posts were categorized as moderate risk (54.34\%), while 15.84\% were classified as high risk and 29.83\% as low risk. These results show that the identified misinformation varies substantially in its potential for harm. Moderate-risk claims may contribute to confusion, treatment hesitancy, or distorted perceptions of screening and care, while high-risk misinformation may more directly encourage treatment avoidance or unsupported interventions. This distinction underscores the importance of evaluating misinformation by potential harm, not occurrence alone~\cite{wang2019systematic}. Detection and moderation systems should therefore account for contextual severity and potential behavioral consequences rather than treating all misinformation as equally harmful.

\noindent\textbf{Implications for Health Communication.}
Our findings show that unproven or alternative treatments constituted the most prevalent misinformation topic, followed by claims about diagnosis and screening and the safety and effectiveness of conventional cancer treatments. This is consistent with prior cancer misinformation research identifying alternative medicine and unsupported treatment claims as recurring concerns~\cite{yussof2023breast,fridman2025identifying}. Our evidence-theme analysis further illustrates the range of specific claims involved, including natural remedies, treatment skepticism, misconceptions about cancer processes, and vaccine-related causation. Such claims may shape treatment choices, screening behaviors, and trust in clinical recommendations, particularly when presented through personal experiences or peer advice. Given the importance of timely, evidence-based cancer care, even seemingly moderate claims may contribute to confusion or treatment hesitancy~\cite{vraga2020correction,pennycook2019understanding}. Public health and clinical communication strategies should address both misconceptions about established cancer care and unsupported alternative approaches by providing accessible evidence while also addressing uncertainty, treatment concerns, and questions about conventional care.
\section{Conclusion} \label{sec:conclusion}
We introduced a multi-dimensional taxonomy for analyzing cancer misinformation on Reddit and constructed an expert-annotated dataset spanning multiple cancer communities. Our results show that LLMs can support scalable, fine-grained misinformation annotation, while performance varies substantially across taxonomy dimensions, with more subjective dimensions such as Risk Level remaining challenging. Applying the framework to more than 133K Reddit posts revealed substantial variation in how misinformation is communicated, engaged with, and associated with potential harm, with unproven or alternative treatments emerging as the most prevalent topic. Our taxonomy, dataset, and analysis provide a foundation for moving beyond binary detection toward multi-dimensional analysis of online cancer misinformation at scale.
\section*{Acknowledgment} \label{sec:acknowledgment}
We thank Joyston Menezes for his assistance and contributions to this work. We thank Google for supporting this research through a Gemini Academic Award. We also thank the Reddit users whose public discussions made this research possible, particularly those who shared personal experiences, questions, and perspectives related to cancer and cancer care.

\section{Limitations} \label{sec:limitation}
This study has several limitations. First, our analysis is based exclusively on Reddit data and therefore may not generalize to the broader population of cancer patients, caregivers, or health-information consumers. Reddit users tend to be younger and more technologically engaged than the general population, further limiting the representativeness of our findings. In addition, cancer-related discussions on Reddit may differ substantially from conversations occurring on other platforms such as Facebook, TikTok, YouTube, or closed patient-support communities, where misinformation dynamics, moderation practices, and interaction structures vary considerably.
Second, our dataset focuses primarily on specialized cancer-related subreddits that are relatively moderated and topic-focused. This focus may underestimate the diversity and extremity of misinformation circulating in more general-purpose, politically polarized, or weakly moderated communities. Future work should extend this analysis to cross-platform and broader community settings to better understand how cancer misinformation spreads across heterogeneous online environments.

Third, although the evaluated LLMs achieved strong overall classification performance, automated annotation may be imperfect. Some misinformation cases involve sarcasm, humor, anecdotal experience, implicit implications, evolving medical evidence, or mixtures of factual and misleading claims that remain challenging even for advanced models. These ambiguities are particularly important in health discussions, where personal experiences may coexist with unsupported recommendations or inaccurate medical interpretations. As a result, some automatically labeled posts may contain classification errors that propagate into downstream analyses. In addition, although we selected more stable versions of proprietary LLMs where possible, provider updates may alter model behavior over time, potentially affecting the reproducibility of model-based annotations.
Fourth, our human-annotated evaluation set reflects a deliberate sampling strategy rather than a random sample of the full corpus. To ensure sufficient representation of misinformation for expert annotation, candidate posts were identified through weak labeling and multi-model consensus, and the annotation set was stratified across consensus-positive and consensus-negative cases. This consensus-enriched sampling may underrepresent ambiguous cases near the misinformation decision boundary. 

Fifth, several taxonomy dimensions involve substantial contextual and subjective judgment. Risk Level showed low agreement under both Cohen's $\kappa$ and Gwet's AC1 and yielded the weakest automated classification performance, while differences between $\kappa$ and AC1 for other dimensions indicate sensitivity to uneven label distributions. Although disagreements were adjudicated by a third expert to obtain final labels, adjudication collapses some underlying uncertainty into a single label. Future work will examine approaches that explicitly model annotator disagreement and uncertainty.
Sixth, the taxonomy itself reflects a particular operationalization of cancer misinformation informed by prior literature and domain expertise. Although we designed the framework to capture multiple dimensions of misinformation, alternative taxonomies or annotation schemes could emphasize different aspects of online health discourse, such as emotional persuasion, uncertainty, scientific evidence quality, or community-level dynamics. Similarly, our topic categories and risk definitions may evolve as medical consensus, online narratives, and misinformation strategies change over time.
Finally, our study focuses on English-language discussions and does not address multilingual or cross-cultural misinformation narratives. Cancer misinformation may vary across healthcare systems and cultural contexts, including around screening, alternative medicine, and institutional trust. Extending multi-dimensional misinformation modeling to multilingual and culturally diverse settings remains an important direction for future research.

\section{Ethical Considerations} \label{sec:ethics}
This work analyzes Reddit discussions related to cancer and health misinformation. Although Reddit data is publicly accessible, discussions about cancer frequently involve sensitive personal experiences, medical histories, treatment decisions, emotional distress, and caregiver concerns. To reduce privacy risks, we removed deleted and removed posts from the dataset and excluded user-identifying information from our analysis and reporting. We do not release usernames, account identifiers, or raw post content containing personally identifiable information, and we will only share Post IDs in our dataset. Example posts included in the paper were paraphrased and anonymized to reduce re-identification risk while preserving the substantive characteristics necessary for scientific interpretation.

Our annotations were produced by domain experts and adjudicated by an experienced oncologist (see \S\ref{sec:annotation_agreement} for details). Medical evidence and its interpretation can also evolve over time. Consequently, our labels should not be interpreted as definitive clinical judgments or medical advice. In particular, some posts classified as misinformation may reflect patient uncertainty, incomplete understanding, personal experience, or evolving scientific evidence rather than intentional deception.
The use of LLMs for large-scale annotation introduces additional ethical considerations. Although our selected models demonstrated strong classification performance, automated systems may encode biases from their training data and may disproportionately misclassify certain linguistic styles, communities, or forms of expression. These biases may influence both misinformation detection and downstream taxonomy analyses. Although we use cross-model agreement to reduce reliance on a single classifier, model agreement does not constitute clinical ground truth and may reproduce biases or blind spots shared across models.

Our work is intended to support research on online health misinformation, not to enable censorship or punitive moderation. Automated classification can incorrectly flag legitimate patient concerns, personal experiences, or good-faith questions about treatment alternatives. This risk is particularly important in health contexts where users seek emotional support, second opinions, or experiential knowledge. Our finding that misinformation frequently appears within questioning, mixed-stance, and information-sharing exchanges further illustrates why a positive classification should not be interpreted as evidence of deliberate misinformation sharing or malicious intent. More broadly, operationalizing health misinformation creates a tension between identifying potentially harmful content and preserving legitimate uncertainty and discussion. We therefore caution against deploying automated systems as standalone moderation tools without human oversight and contextual review. Our multi-dimensional framework is intended in part to preserve these distinctions rather than reducing health discourse to binary true/false judgments.

\bibliography{custom}

@inproceedings{rajwal2024unveiling,
  title={Unveiling voices: Identification of concerns in a social media breast cancer cohort via natural language processing},
  author={Rajwal, Swati and Pandey, Avinash Kumar and Han, Zhishuo and Sarker, Abeed},
  booktitle={Proceedings of the First Workshop on Patient-Oriented Language Processing (CL4Health)@ LREC-COLING 2024},
  url= "https://aclanthology.org/2024.cl4health-1.32/",
  pages={264--270},
  year={2024}
}

@inproceedings{podina-etal-2021-natural,
    title = "Natural language processing as a tool to identify the {R}eddit particularities of cancer survivors around the time of diagnosis and remission: A pilot study",
    author = "Podin{\u{a}}, Ioana R  and
      Bucur, Ana-Maria  and
      Todea, Diana  and
      Fodor, Liviu  and
      Luca, Andreea  and
      Dinu, Liviu P.  and
      Boian, Rareș",
    booktitle = "Proceedings of the Fifth Workshop on Widening Natural Language Processing",
    month = nov,
    year = "2021",
    address = "Punta Cana, Dominican Republic",
    publisher = "Association for Computational Linguistics",
    url = "https://aclanthology.org/2021.winlp-1.4/",
    pages = "21--24",
}

@inproceedings{liu2024breaking,
  title={Breaking the Silence: How Online Forums Address Lung Cancer Stigma and Offer Support},
  author={Liu, Jiahe and Conway, Mike and Lozoya, Daniel Cabrera},
  booktitle={Proceedings of the 22nd Annual Workshop of the Australasian Language Technology Association},
  pages={179--188},
  year={2024},
  publisher={Association for Computational Linguistics},
  url={https://aclanthology.org/2024.alta-1.15/}
}

@inproceedings{bouzoubaa2023exploring,
  title={Exploring the landscape of drug communities on reddit: A network study},
  author={Bouzoubaa, Layla and Young, Jordyn and Rezapour, Rezvaneh},
  booktitle={Proceedings of the International Conference on Advances in Social Networks Analysis and Mining},
  pages={558--565},
  year={2023},
  publisher={Association for Computing Machinery},
  doi={10.1145/3625007.3629125},
  url={https://dl.acm.org/doi/10.1145/3625007.3629125}
}

@inproceedings{lal2024analysing,
  title={Analysing emotions in cancer narratives: a corpus-driven approach},
  author={Lal, Daisy Monika and Rayson, Paul and Payne, Sheila A and Liu, Yufeng},
  booktitle={Proceedings of the First Workshop on Patient-Oriented Language Processing (CL4Health)@ LREC-COLING 2024},
  pages={73--83},
  publisher={ELRA and ICCL},
  url= "https://aclanthology.org/2024.cl4health-1.9/",
  year={2024}
}

@inproceedings{lal2025lens,
  title={{LENS}: Learning Entities from Narratives of Skin Cancer},
  author={Lal, Daisy Monika and Rayson, Paul and Peter, Christopher and Ezeani, Ignatius and El-Haj, Mo and Zhu, Yafei and Liu, Yufeng},
  booktitle={Proceedings of the 31st International Conference on Computational Linguistics: System Demonstrations},
  pages={20--27},
  year={2025},
  publisher={Association for Computational Linguistics},
  url={https://aclanthology.org/2025.coling-demos.3/}
}

@article{swire2024cancer,
  title={Cancer: A model topic for misinformation researchers},
  author={Swire-Thompson, Briony and Johnson, Skyler},
  journal={Current Opinion in Psychology},
  volume={56},
  pages={101775},
  year={2024},
  publisher={Elsevier},
  doi={10.1016/j.copsyc.2023.101775},
  url={https://doi.org/10.1016/j.copsyc.2023.101775}
}

@article{loeb2024cancer,
  title={Cancer misinformation on social media},
  author={Loeb, Stacy and Langford, Aisha T and Bragg, Marie A and Sherman, Robert and Chan, June M},
  journal={CA: A Cancer Journal for Clinicians},
  volume={74},
  number={5},
  pages={453--464},
  year={2024},
  publisher={Wiley Online Library},
  doi={10.3322/caac.21857},
  url={https://doi.org/10.3322/caac.21857}
}

@article{pierce2024identifying,
  title={Identifying factors associated with heightened anxiety during breast cancer diagnosis through the analysis of social media data on {Reddit}: Mixed methods study},
  author={Pierce, Joni and Conway, Mike and Grace, Kathryn and Mikal, Jude},
  journal={JMIR cancer},
  volume={10},
  number={1},
  pages={e52551},
  year={2024},
  publisher={JMIR Publications Inc., Toronto, Canada},
  doi={10.2196/52551},
  url={https://doi.org/10.2196/52551}
}

@article{yussof2023breast,
  title={Breast cancer prevention and treatment misinformation on {Twitter}: an analysis of two languages},
  author={Yussof, Izzati and Ab Muin, Nur Fa'izah and Mohd, Masnizah and Hatah, Ernieda and Mohd Tahir, Nor Asyikin and Mohamed Shah, Noraida},
  journal={Digital Health},
  volume={9},
  pages={20552076231205742},
  year={2023},
  doi={10.1177/20552076231205742},
  url={https://doi.org/10.1177/20552076231205742},
  publisher={SAGE Publications Sage UK: London, England}
}

@article{fridman2025identifying,
  title={Identifying misinformation about unproven cancer treatments on social media using user-friendly linguistic characteristics: content analysis},
  author={Fridman, Ilona and Boyles, Dahlia and Chheda, Ria and Baldwin-SoRelle, Carrie and Smith, Angela B and Lafata, Jennifer Elston},
  journal={JMIR infodemiology},
  url= "https://infodemiology.jmir.org/2025/1/e62703/",
  volume={5},
  number={1},
  pages={e62703},
  year={2025},
  publisher={JMIR Publications Inc., Toronto, Canada}
}

@article{lazard2026intervening,
  title={Intervening and reducing sharing of false cancer treatments on social media: Online experiment},
  author={Lazard, Allison J and Lake, Shelby and Queen, Tara Licciardello and Avenda{\~n}o-Galdamez, Mirian and Babwah Brennen, Scott and Varma, Tushar and Tan, Hung-Jui and Charlot, Marjory and Dasgupta, Nabarun},
  journal={PloS one},
  volume={21},
  number={2},
  pages={e0341907},
  year={2026},
  publisher={Public Library of Science San Francisco, CA USA},
  doi={10.1371/journal.pone.0341907},
  url={https://doi.org/10.1371/journal.pone.0341907}
}

@article{sehat2024misinformation,
  title={Misinformation as a harm: structured approaches for fact-checking prioritization},
  author={Sehat, Connie Moon and Li, Ryan and Nie, Peipei and Prabhakar, Tarunima and Zhang, Amy X},
  journal={Proceedings of the ACM on human-computer interaction},
  volume={8},
  number={CSCW1},
  pages={1--36},
  year={2024},
  publisher={ACM New York, NY, USA},
  doi={10.1145/3641010},
  url={https://doi.org/10.1145/3641010}
}

@inproceedings{tran2019investigation,
  title={An investigation of misinformation harms related to social media during humanitarian crises},
  author={Tran, Thi and Valecha, Rohit and Rad, Paul and Rao, H Raghav},
  booktitle={International conference on secure knowledge management in artificial intelligence era},
  pages={167--181},
  year={2019},
  organization={Springer},
  doi={10.1007/978-981-15-3817-9_10},
  url={https://doi.org/10.1007/978-981-15-3817-9_10}
}

@article{tran2020misinformation,
  title={Misinformation harms: A tale of two humanitarian crises},
  author={Tran, Thi and Valecha, Rohit and Rad, Paul and Rao, H Raghav},
  journal={IEEE Transactions on Professional Communication},
  volume={63},
  number={4},
  pages={386--399},
  year={2020},
  publisher={IEEE},
  doi={10.1109/TPC.2020.3029685},
  url={https://doi.org/10.1109/TPC.2020.3029685}
}

@article{alvarez2025development,
  title={Development of a Conceptual Framework of Health Misinformation During the {COVID-19} Pandemic: Systematic Review of Reviews},
  author={Alvarez-Galvez, Javier and Carretero-Bravo, Jesus and Lagares-Franco, Carolina and Ramos-Fiol, Bego{\~n}a and Ortega-Martin, Esther},
  journal={JMIR Public Health and Surveillance},
  url= "https://publichealth.jmir.org/2025/1/e62693/",
  volume={11},
  number={1},
  pages={e62693},
  year={2025},
  publisher={JMIR Publications Inc., Toronto, Canada}
}

@article{warner2021young,
  title={Young adult cancer caregivers' exposure to cancer misinformation on social media},
  author={Warner, Echo L and Waters, Austin R and Cloyes, Kristin G and Ellington, Lee and Kirchhoff, Anne C},
  journal={Cancer},
  volume={127},
  number={8},
  pages={1318--1324},
  year={2021},
  publisher={Wiley Online Library},
  doi={10.1002/cncr.33380},
  url={https://doi.org/10.1002/cncr.33380}
}

@article{johnson2022cancer,
  title={Cancer misinformation and harmful information on {Facebook} and other social media: a brief report},
  author={Johnson, Skyler B and Parsons, Matthew and Dorff, Tanya and Moran, Meena S and Ward, John H and Cohen, Stacey A and Akerley, Wallace and Bauman, Jessica and Hubbard, Joleen and Spratt, Daniel E and others},
  journal={JNCI: Journal of the National Cancer Institute},
  volume={114},
  number={7},
  pages={1036--1039},
  year={2022},
  doi={10.1093/jnci/djab141},
  url={https://doi.org/10.1093/jnci/djab141},
  publisher={Oxford University Press}
}

@misc{Baumgartner2020Pushshift,
      title={The Pushshift {Reddit} Dataset}, 
      author={Jason Baumgartner and Savvas Zannettou and Brian Keegan and Megan Squire and Jeremy Blackburn},
      booktitle={Proceedings of the International AAAI Conference on Web and Social Media},
      volume={14},
      number={1},
      pages={830--839},
      year={2020},
      publisher={Association for the Advancement of Artificial Intelligence},
      doi={10.1609/icwsm.v14i1.7347},
      url={https://ojs.aaai.org/index.php/ICWSM/article/view/7347}
}

@misc{Heitmann2025arctic_shift,
  author       = {Arthur Heitmann},
  title        = {arctic\_shift: Making {Reddit} data accessible to researchers, moderators and everyone else},
  year         = {2025},
  url          = {https://github.com/ArthurHeitmann/arctic_shift},
  note         = {Accessed: 2025-07-15}
}

@article{lazard2023exposure,
  title={Exposure and reactions to cancer treatment misinformation and advice: survey study},
  author={Lazard, Allison J and Nicolla, Sydney and Vereen, Rhyan N and Pendleton, Shanetta and Charlot, Marjory and Tan, Hung-Jui and DiFranzo, Dominic and Pulido, Marlyn and Dasgupta, Nabarun},
  journal={JMIR cancer},
  volume={9},
  pages={e43749},
  year={2023},
  publisher={JMIR Publications Toronto, Canada},
  doi={10.2196/43749},
  url={https://doi.org/10.2196/43749}
}

@inproceedings{de2014mental,
  title={Mental health discourse on {Reddit}: Self-disclosure, social support, and anonymity},
  author={De Choudhury, Munmun and De, Sushovan},
  booktitle={Proceedings of the international AAAI conference on web and social media},
  volume={8},
  number={1},
  pages={71--80},
  year={2014},
  publisher={Association for the Advancement of Artificial Intelligence},
  doi={10.1609/icwsm.v8i1.14526},
  url={https://ojs.aaai.org/index.php/ICWSM/article/view/14526}
}

@article{sarker2015utilizing,
  title={Utilizing social media data for pharmacovigilance: a review},
  author={Sarker, Abeed and Ginn, Rachel and Nikfarjam, Azadeh and O'Connor, Karen and Smith, Karen and Jayaraman, Swetha and Upadhaya, Tejaswi and Gonzalez, Graciela},
  journal={Journal of biomedical informatics},
  volume={54},
  pages={202--212},
  year={2015},
  publisher={Elsevier},
  doi={10.1016/j.jbi.2015.02.004},
  url={https://doi.org/10.1016/j.jbi.2015.02.004}
}

@book{wardle2017information,
  title={Information disorder: Toward an interdisciplinary framework for research and policymaking},
  author={Wardle, Claire and Derakhshan, Hossein},
  volume={27},
  year={2017},
  publisher={Council of Europe Strasbourg},
  url={https://rm.coe.int/information-disorder-toward-an-interdisciplinary-framework-for-researc/168076277c}
}

@article{nagler2014adverse,
  title={Adverse outcomes associated with media exposure to contradictory nutrition messages},
  author={Nagler, Rebekah H},
  journal={Journal of health communication},
  volume={19},
  number={1},
  pages={24--40},
  year={2014},
  publisher={Taylor \& Francis},
  doi={10.1080/10810730.2013.798384},
  url={https://doi.org/10.1080/10810730.2013.798384}
}

@article{chou2018addressing,
  title={Addressing health-related misinformation on social media},
  author={Chou, Wen-Ying Sylvia and Oh, April and Klein, William MP},
  journal={Jama},
  doi={10.1001/jama.2018.16865},
  url={https://doi.org/10.1001/jama.2018.16865},
  volume={320},
  number={23},
  pages={2417--2418},
  year={2018}
}

@article{lambert2007health,
  title={Health information—seeking behavior},
  author={Lambert, Sylvie D and Loiselle, Carmen G},
  journal={Qualitative health research},
  volume={17},
  number={8},
  pages={1006--1019},
  year={2007},
  publisher={Sage Publications Sage CA: Los Angeles, CA},
  doi={10.1177/1049732307305199},
  url={https://doi.org/10.1177/1049732307305199}
}

@article{southwell2019misinformation,
  title={Misinformation as a misunderstood challenge to public health},
  author={Southwell, Brian G and Niederdeppe, Jeff and Cappella, Joseph N and Gaysynsky, Anna and Kelley, Dannielle E and Oh, April and Peterson, Emily B and Chou, Wen-Ying Sylvia},
  journal={American journal of preventive medicine},
  volume={57},
  number={2},
  pages={282--285},
  year={2019},
  publisher={Elsevier},
  doi={10.1016/j.amepre.2019.03.009},
  url={https://doi.org/10.1016/j.amepre.2019.03.009}
}

@inproceedings{pei2022potato,
  title={{POTATO}: The Portable Text Annotation Tool},
  author={Pei, Jiaxin and Ananthasubramaniam, Aparna and Wang, Xingyao and Zhou, Naitian and Dedeloudis, Apostolos and Sargent, Jackson and Jurgens, David},
  booktitle={Proceedings of the 2022 Conference on Empirical Methods in Natural Language Processing: System Demonstrations},
  pages={327--337},
  year={2022},
  publisher={Association for Computational Linguistics},
  doi={10.18653/v1/2022.emnlp-demos.33},
  url={https://aclanthology.org/2022.emnlp-demos.33/}
  
}

@article{wei2022chain,
  title={Chain-of-thought prompting elicits reasoning in large language models},
  author={Wei, Jason and Wang, Xuezhi and Schuurmans, Dale and Bosma, Maarten and Xia, Fei and Chi, Ed and Le, Quoc V and Zhou, Denny and others},
  journal={Advances in neural information processing systems},
  volume={35},
  pages={24824--24837},
  year={2022},
  url={https://arxiv.org/pdf/2201.11903}
}

@article{warner2022online,
  title={The Online Cancer Nutrition Misinformation: A framework of behavior change based on exposure to cancer nutrition misinformation},
  author={Warner, Echo L and Basen-Engquist, Karen M and Badger, Terry A and Crane, Tracy E and Raber-Ramsey, Margaret},
  journal={Cancer},
  volume={128},
  number={13},
  pages={2540--2548},
  year={2022},
  publisher={Wiley Online Library},
  doi={10.1002/cncr.34218},
  url={https://doi.org/10.1002/cncr.34218}
}

@article{wilner2020breast,
  title={Breast cancer prevention and treatment: misinformation on {Pinterest}, 2018},
  author={Wilner, Tamar and Holton, Avery},
  journal={American journal of public health},
  volume={110},
  number={S3},
  pages={S300--S304},
  year={2020},
  publisher={American Public Health Association},
  doi={10.2105/AJPH.2020.305812},
  url={https://doi.org/10.2105/AJPH.2020.305812}
}

@article{trivedi2022factors,
  title={Factors associated with cancer message believability: a mixed methods study on simulated {Facebook} posts},
  author={Trivedi, Neha and Lowry, Mark and Gaysynsky, Anna and Chou, Wen-Ying Sylvia},
  journal={Journal of Cancer Education},
  volume={37},
  number={6},
  pages={1870--1878},
  year={2022},
  publisher={Springer},
  doi={10.1007/s13187-021-02054-7},
  url={https://doi.org/10.1007/s13187-021-02054-7}
}

@article{varet2025assessing,
  title={Assessing the role of conspiracy beliefs in oncological treatment decisions: An experimental approach},
  author={Varet, Florent and Fournier, Valentyn and Delouv{\'e}e, Sylvain},
  journal={Applied Psychology: Health and Well-Being},
  volume={17},
  number={1},
  pages={e12615},
  year={2025},
  publisher={Wiley Online Library},
  doi={10.1111/aphw.12615},
  url={https://doi.org/10.1111/aphw.12615}
}

@article{kata2010postmodern,
  title={A postmodern {Pandora's} box: anti-vaccination misinformation on the Internet},
  author={Kata, Anna},
  journal={Vaccine},
  volume={28},
  number={7},
  pages={1709--1716},
  year={2010},
  publisher={Elsevier},
  doi={10.1016/j.vaccine.2009.12.022},
  url={https://doi.org/10.1016/j.vaccine.2009.12.022}
}

@inproceedings{pessianzadeh2025exploring,
  title={Exploring stance on affirmative action through {Reddit} narratives},
  author={Pessianzadeh, Aria and Rezapour, Rezvaneh},
  booktitle={Proceedings of the 17th ACM Web Science Conference 2025},
  pages={52--63},
  year={2025},
  publisher={Association for Computing Machinery},
  doi={10.1145/3717867.3717891},
  url={https://doi.org/10.1145/3717867.3717891}
}

@article{meteran2023usefulness,
  title={The usefulness of {YouTube} videos on lung cancer},
  author={Meteran, Hanieh and H{\o}j, Simon and Sigsgaard, Torben and Diers, Caroline Skovsgaard and Remvig, Celine and Meteran, Howraman},
  journal={Journal of Public Health},
  volume={45},
  number={2},
  pages={e339--e345},
  year={2023},
  publisher={Oxford University Press},
  doi={10.1093/pubmed/fdac092},
  url={https://doi.org/10.1093/pubmed/fdac092}
}

@article{komsany2026leveraging,
  title={Leveraging {AI} for Analysis of Digital Health Information on Cancer Prevention Among {Arab} Youth and Adults: Content Analysis},
  author={Komsany, Alia and Al Zoubi, Obada and Sebaaly, Laetitia and Harrison, Gabrielle and Soroka, Orysya and ElKefi, Safa and Scales, David and Phillips, Erica and Pinheiro, Laura C and Ismail, Israa and others},
  journal={JMIR infodemiology},
  volume={6},
  pages={e77888},
  year={2026},
  url= "https://infodemiology.jmir.org/2026/1/e77888",
  publisher={JMIR Publications Toronto, Canada}
}

@article{peng2023persuasive,
  title={Persuasive strategies in online health misinformation: a systematic review},
  author={Peng, Wei and Lim, Sue and Meng, Jingbo},
  journal={Information, Communication \& Society},
  volume={26},
  number={11},
  pages={2131--2148},
  year={2023},
  publisher={Taylor \& Francis},
  doi={10.1080/1369118X.2022.2085615},
  url={https://doi.org/10.1080/1369118X.2022.2085615}
}

@inproceedings{kamali2024using,
  title={Using persuasive writing strategies to explain and detect health misinformation},
  author={Kamali, Danial and Romain, Joseph D and Liu, Huiyi and Peng, Wei and Meng, Jingbo and Kordjamshidi, Parisa},
  booktitle={Proceedings of the 2024 Joint International Conference on Computational Linguistics, Language Resources and Evaluation (LREC-COLING 2024)},
  pages={17285--17309},
  url= "https://aclanthology.org/2024.lrec-main.1501/",
  year={2024}
}

@article{baker2025conspiracy,
  title={Conspiracy theory as metaphor: Promoting cancer misinformation through algorithmic influence and metaphoric manipulation on {TikTok}},
  author={Baker, Stephanie Alice},
  journal={Journal of Religion, Media and Digital Culture},
  url= "https://brill.com/view/journals/rmdc/14/2/article-p193_003.xml?language=en&srsltid=AfmBOopcudjjRszgc4hyDgCAKUx4bnIuCrd1yzX03Xt0NUzobAxKe998",
  volume={14},
  number={2},
  pages={193--215},
  year={2025},
  publisher={Brill}
}

@article{brown2020language,
  title={Language models are few-shot learners},
  author={Brown, Tom and Mann, Benjamin and Ryder, Nick and Subbiah, Melanie and Kaplan, Jared D and Dhariwal, Prafulla and Neelakantan, Arvind and Shyam, Pranav and Sastry, Girish and Askell, Amanda and others},
  journal={Advances in neural information processing systems},
  url= "https://proceedings.neurips.cc/paper_files/paper/2020/file/1457c0d6bfcb4967418bfb8ac142f64a-Paper.pdf",
  volume={33},
  pages={1877--1901},
  year={2020}
}

@article{kuccuk2020stance,
  title={Stance detection: A survey},
  author={K{\"u}{\c{c}}{\"u}k, Dilek and Can, Fazli},
  journal={ACM Computing Surveys (CSUR)},
  volume={53},
  number={1},
  pages={1--37},
  year={2020},
  doi={10.1145/3369026},
  url={https://doi.org/10.1145/3369026},
  publisher={ACM New York, NY, USA}
}

@inproceedings{rezapour2021incorporating,
  title={Incorporating the measurement of moral foundations theory into analyzing stances on controversial topics},
  author={Rezapour, Rezvaneh and Dinh, Ly and Diesner, Jana},
  booktitle={Proceedings of the 32nd ACM conference on hypertext and social media},
  pages={177--188},
  year={2021},
  doi={10.1145/3465336.3475112},
  url={https://doi.org/10.1145/3465336.3475112}
}

@inproceedings{lam2024conceptInduction,
    author = {Lam, Michelle S. and Teoh, Janice and Landay, James and Heer, Jeffrey and Bernstein, Michael S.},
    title = {Concept Induction: Analyzing Unstructured Text with High-Level Concepts Using LLooM},
    year = {2024},
    isbn = {9798400703300},
    publisher = {Association for Computing Machinery},
    address = {New York, NY, USA},
    url = {https://doi.org/10.1145/3613904.3642830},
    doi = {10.1145/3613904.3642830},
    booktitle = {Proceedings of the 2024 CHI Conference on Human Factors in Computing Systems},
    articleno = {933},
    numpages = {28},
    location = {Honolulu, HI, USA},
    series = {CHI '24}
}

@inproceedings{jia2021online,
  title={Online health information seeking behavior: a systematic review},
  author={Jia, Xiaoyun and Pang, Yan and Liu, Liangni Sally},
  booktitle={Healthcare},
  number={12},
  pages={1740},
  year={2021},
  doi={10.3390/healthcare9121740},
  url= "https://www.mdpi.com/2227-9032/9/12/1740",
  organization={MDPI}
}

@article{moorhead2013new,
  title={A new dimension of health care: systematic review of the uses, benefits, and limitations of social media for health communication},
  author={Moorhead, S Anne and Hazlett, Diane E and Harrison, Laura and Carroll, Jennifer K and Irwin, Anthea and Hoving, Ciska},
  journal={Journal of medical Internet research},
  volume={15},
  number={4},
  pages={e1933},
  year={2013},
  publisher={JMIR Publications Inc., Toronto, Canada},
  doi={10.2196/jmir.1933},
  url={https://doi.org/10.2196/jmir.1933}
}

@article{wang2019systematic,
  title={Systematic literature review on the spread of health-related misinformation on social media},
  author={Wang, Yuxi and McKee, Martin and Torbica, Aleksandra and Stuckler, David},
  journal={Social science \& medicine},
  volume={240},
  pages={112552},
  year={2019},
  publisher={Elsevier},
  doi={10.1016/j.socscimed.2019.112552},
  url={https://doi.org/10.1016/j.socscimed.2019.112552}
}

@article{wu2022linking,
  title={Linking social media overload to health misinformation dissemination: An investigation of the underlying mechanisms},
  author={Wu, Manli and Pei, Yiming},
  journal={Telematics and Informatics Reports},
  volume={8},
  pages={100020},
  year={2022},
  publisher={Elsevier},
  doi={10.1016/j.teler.2022.100020},
  url={https://doi.org/10.1016/j.teler.2022.100020}
}

@article{team2023gemini,
  title={Gemini: a family of highly capable multimodal models},
  author={Team, Gemini and Anil, Rohan and Borgeaud, Sebastian and Alayrac, Jean-Baptiste and Yu, Jiahui and Soricut, Radu and Schalkwyk, Johan and Dai, Andrew M and Hauth, Anja and Millican, Katie and others},
  journal={arXiv preprint arXiv:2312.11805},
  year={2023},
  doi={10.48550/arXiv.2312.11805},
  url={https://arxiv.org/abs/2312.11805}
}

@article{yang2025qwen3,
  title={Qwen3 technical report},
  author={Yang, An and Li, Anfeng and Yang, Baosong and Zhang, Beichen and Hui, Binyuan and Zheng, Bo and Yu, Bowen and Gao, Chang and Huang, Chengen and Lv, Chenxu and others},
  journal={arXiv preprint arXiv:2505.09388},
  year={2025},
  doi={10.48550/arXiv.2505.09388},
  url={https://arxiv.org/abs/2505.09388}
}

@article{grattafiori2024llama,
  title={The llama 3 herd of models},
  author={Grattafiori, Aaron and Dubey, Abhimanyu and Jauhri, Abhinav and Pandey, Abhinav and Kadian, Abhishek and Al-Dahle, Ahmad and Letman, Aiesha and Mathur, Akhil and Schelten, Alan and Vaughan, Alex and others},
  journal={arXiv preprint arXiv:2407.21783},
  year={2024},
  doi={10.48550/arXiv.2407.21783},
  url={https://doi.org/10.48550/arXiv.2407.21783}
}

@misc{anthropic2026claude,
  author       = {{Anthropic}},
  title        = {Claude 3.5 Sonnet},
  howpublished = {Large language model},
  year         = {2026},
  url          = {https://www.anthropic.com}
}

@article{team2024gemma,
  title={Gemma: Open models based on gemini research and technology},
  author={Team, Gemma and Mesnard, Thomas and Hardin, Cassidy and Dadashi, Robert and Bhupatiraju, Surya and Pathak, Shreya and Sifre, Laurent and Rivi{\`e}re, Morgane and Kale, Mihir Sanjay and Love, Juliette and others},
  journal={arXiv preprint arXiv:2403.08295},
  year={2024},
  doi={10.48550/arXiv.2403.08295},
  url={https://arxiv.org/abs/2403.08295}
}

@article{podina2023mental,
  title={Mental health at different stages of cancer survival: a natural language processing study of Reddit posts},
  author={Podin{\u{a}}, Ioana R and Bucur, Ana-Maria and Todea, Diana and Fodor, Liviu and Luca, Andreea and Dinu, Liviu P and Boian, Rareș F},
  journal={Frontiers in Psychology},
  volume={14},
  pages={1150227},
  year={2023},
  doi={10.3389/fpsyg.2023.1150227},
  url={https://doi.org/10.3389/fpsyg.2023.1150227},
  publisher={Frontiers Media SA}
}

@article{suarez2021prevalence,
  title={Prevalence of health misinformation on social media: systematic review},
  author={Suarez-Lledo, Victor and Alvarez-Galvez, Javier},
  journal={Journal of medical Internet research},
  volume={23},
  number={1},
  pages={e17187},
  year={2021},
  publisher={JMIR Publications Toronto, Canada},
  doi={10.2196/17187},
  url={https://doi.org/10.2196/17187}
}

@article{broniatowski2020twitter,
  title={Twitter and Facebook posts about COVID-19 are less likely to spread false and low-credibility content compared to other health topics},
  author={Broniatowski, David A and Kerchner, Daniel and Farooq, Fouzia and Huang, Xiaolei and Jamison, Amelia M and Dredze, Mark and Quinn, Sandra Crouse},
  journal={PLOS ONE},
  volume={17},
  number={1},
  pages={e0261768},
  year={2022},
  publisher={Public Library of Science},
  doi={10.1371/journal.pone.0261768},
  url={https://doi.org/10.1371/journal.pone.0261768}
}

@article{van2022misinformation,
  title={Misinformation: susceptibility, spread, and interventions to immunize the public},
  author={Van Der Linden, Sander},
  journal={Nature medicine},
  volume={28},
  number={3},
  pages={460--467},
  year={2022},
  publisher={Nature Publishing Group US New York},
  doi={10.1038/s41591-022-01713-6},
  url={https://doi.org/10.1038/s41591-022-01713-6}
}

@article{cinelli2020covid,
  title={The COVID-19 social media infodemic},
  author={Cinelli, Matteo and Quattrociocchi, Walter and Galeazzi, Alessandro and Valensise, Carlo Michele and Brugnoli, Emanuele and Schmidt, Ana Lucia and Zola, Paola and Zollo, Fabiana and Scala, Antonio},
  journal={Scientific reports},
  url= "https://www.nature.com/articles/s41598-020-73510-5",
  volume={10},
  number={1},
  pages={16598},
  year={2020},
  publisher={Springer Nature}
}

@misc{vraga2020correction,
  title={Correction as a solution for health misinformation on social media},
  author={Vraga, Emily K and Bode, Leticia},
  journal={American Journal of Public Health},
  volume={110},
  number={S3},
  pages={S278--S280},
  year={2020},
  publisher={American Public Health Association},
  doi={10.2105/AJPH.2020.305916},
  url={https://doi.org/10.2105/AJPH.2020.305916}
}

@article{pennycook2019understanding,
  title={Understanding and reducing the spread of misinformation online},
  author={Pennycook, Gordon and Epstein, Ziv and Mosleh, Mohsen and Arechar, Antonio A and Eckles, Dean and Rand, David G},
  journal={Unpublished manuscript: https://psyarxiv. com/3n9u8},
  pages={1--84},
  year={2019},
}

@inproceedings{pessianzadeh2026reddit,
  title={Reddit After {Roe}: A Computational Analysis of Abortion Narratives and Barriers in the Wake of {Dobbs}},
  author={Pessianzadeh, Aria and Poole, Alex H and Rezapour, Rezvaneh},
  booktitle={Proceedings of the International AAAI Conference on Web and Social Media},
  volume={20},
  number={1},
  pages={1829--1848},
  year={2026},
  url={https://doi.org/10.1609/icwsm.v20i1.42725}
}

@inproceedings{phadke2026reinforcing,
  title={Reinforcing the Unreal: Subliminals and the Normalization of Unscientific Body Transformations on Reddit},
  author={Phadke, Shruti},
  booktitle={Proceedings of the 2026 CHI Conference on Human Factors in Computing Systems},
  pages={1--19},
  year={2026},
  publisher={Association for Computing Machinery},
  doi={10.1145/3772318.3790846},
  url={https://doi.org/10.1145/3772318.3790846}
}

@inproceedings{aghakhani-rezapour-2026-like,
    title = "Like a Therapist, But Not: {R}eddit Narratives of {AI} in Mental Health Contexts",
    author = "Aghakhani, Elham  and
      Rezapour, Rezvaneh",
    editor = "Liakata, Maria  and
      Moreira, Viviane P.  and
      Zhang, Jiajun  and
      Jurgens, David",
    booktitle = "Findings of the {A}ssociation for {C}omputational {L}inguistics: {ACL} 2026",
    month = jul,
    year = "2026",
    address = "San Diego, California, United States",
    publisher = "Association for Computational Linguistics",
    url = "https://aclanthology.org/2026.findings-acl.569/",
    doi = "10.18653/v1/2026.findings-acl.569",
    pages = "11716--11736",
    ISBN = "979-8-89176-395-1"
}

@article{ghenai2018fake,
  title={Fake cures: user-centric modeling of health misinformation in social media},
  author={Ghenai, Amira and Mejova, Yelena},
  journal={Proceedings of the ACM on human-computer interaction},
  volume={2},
  number={CSCW},
  pages={1--20},
  year={2018},
  url= "https://dl.acm.org/doi/abs/10.1145/3274327",
  publisher={ACM New York, NY, USA}
}

@article{harkin2020secret,
  title={Secret groups and open forums: Defining online support communities from the perspective of people affected by cancer},
  author={Harkin, Lydia Jo and Beaver, Kinta and Dey, Paola and Choong, Kartina Aisha},
  journal={Digital Health},
  volume={6},
  pages={2055207619898993},
  year={2020},
  publisher={SAGE Publications Sage UK: London, England},
  doi={10.1177/2055207619898993},
  url={https://doi.org/10.1177/2055207619898993}
}

@article{tripathi2022examination,
  title={An examination of patients and caregivers on Reddit navigating brain cancer: content analysis of the brain tumor subreddit},
  author={Tripathi, Sanidhya D and Parker, Pearman D and Prabhu, Arpan V and Thomas, Kevin and Rodriguez, Analiz},
  journal={JMIR cancer},
  volume={8},
  number={2},
  pages={e35324},
  year={2022},
  publisher={JMIR Publications Inc., Toronto, Canada},
  doi={10.2196/35324},
  url={https://doi.org/10.2196/35324}
}

@online{wardle2017fakenews,
  author  = {Wardle, Claire},
  title   = {Fake News. It’s Complicated},
  year    = {2017},
  month   = feb,
  day     = {16},
  organization = {First Draft},
  url     = {https://firstdraftnews.org/articles/fake-news-complicated/},
  urldate = {2026-07-14}
}

\renewcommand{\thefigure}{A\arabic{figure}}
\renewcommand{\thetable}{A\arabic{table}}
\setcounter{figure}{0}
\setcounter{table}{0}

\appendix

\section{Cancer-Related Subreddits}
Table~\ref{tab:subreddits list} presents the subreddits selected for each of the four cancer types included in this study: breast, lung, prostate, and colon cancer. These communities were selected based on their relevance to cancer-related discussions.

\begin{table}[h]
\centering
\resizebox{\columnwidth}{!}{%
\begin{tabular}{
    p{0.22\columnwidth}
    p{0.72\columnwidth}
}
\toprule
\textbf{Cancer Type} & \textbf{Subreddits} \\
\midrule

Breast Cancer &
r/breastcancer, r/BreastCancerSurvivors, 
r/BreastCancerResearch, r/BreastCancerGenomics, 
r/doihavebreastcancer, r/LivingWithMBC, 
r/QueerBreastCancer, r/mtf\_breast\_cancer \\[2pt]

Lung Cancer &
r/lungcancer, r/LungCancerSupport, r/nsclc, 
r/SmallCellLungCancer \\[2pt]

Prostate Cancer &
r/ProstateCancer, r/Prostate\_Cancer, 
r/ProstateTreatment, r/ProstateCancer\_AS, 
r/prostateinfo, r/ProstateBob, 
r/ProstateCancerDesign, r/Prostatitis \\[2pt]

Colon/ Colorectal Cancer &
r/colorectalcancer, r/coloncancer, r/colonoscopy, 
r/askcoloncancerdoctors \\

\bottomrule
\end{tabular}%
}
\caption{Cancer-related subreddits included in the study, grouped by cancer type.}
\label{tab:subreddits list}
\end{table}

\section{Data Annotation and Misinformation Identification}
\label{sec:appendix_annotation}

\subsection{Weak-Labeling and Human Annotation Pipeline}
\label{sec:weak_labeling}

Identifying misinformation in the full corpus is challenging because misinformation constitutes a relatively sparse class within a large
collection of predominantly non-misinformation posts. Rather than randomly sampling posts for human annotation, which could yield relatively few positive examples, we used a multi-stage weak-labeling procedure to construct an annotation set enriched for candidate misinformation while retaining a comparison set of likely non-misinformation posts. We first applied Qwen as a weak labeler to the full corpus of \corpusOne{} posts. Qwen classified 1,813 posts as candidate misinformation. We additionally sampled 1,813 posts from those classified as non-misinformation, producing candidate pools of equal size for the two weak-label classes. To increase confidence in the weak labels used for human-sample selection, we independently classified the candidate posts using three additional LLMs: GPT-5.4, GPT-5.2, and Claude Haiku. Among the 1,813 Qwen-positive candidates, 369 posts were also classified as misinformation by all three additional models. From these consensus-positive cases, we sampled 200 posts for human annotation. We similarly selected 200 consensus-negative posts for which the models agreed on the non-misinformation label. This procedure produced a stratified human-annotation set of 400 posts, intentionally enriched for both positive and negative cases rather than designed to reproduce the natural prevalence of misinformation in the full corpus.
Table~\ref{tab:data_attrition} summarizes the annotation and classification pipeline.

\begin{table}[h]
\centering
\small
\resizebox{\columnwidth}{!}{%
\begin{tabular}{
    p{0.25\columnwidth}
    p{0.50\columnwidth}
    r
}
\toprule
\textbf{Stage} & \textbf{Selection mechanism} & \textbf{$N$} \\
\midrule

Full corpus
    & Unlabelled posts
    & \corpusOne{} \\

Qwen candidate positives
    & Posts weakly classified as misinformation
    & 1,813 \\

Qwen comparison negatives
    & Sampled from posts weakly classified as non-misinformation
    & 1,813 \\

Consensus-positive pool
    & Qwen-positive posts also classified as misinformation by
      GPT-5.4, GPT-5.2, and Claude Haiku
    & 369 \\

Human annotation set
    & 200 consensus-positive and 200 consensus-negative posts
    & 400 \\

\bottomrule
\end{tabular}%
}
\caption{Multi-stage weak-labeling and annotation pipeline.}
\label{tab:data_attrition}
\end{table}

\subsection{Human Validation of Weak Labels}
\label{sec:weak_label_validation}

The 400 selected posts were manually annotated according to the study's misinformation annotation criteria. Agreement between the weak-label assignments used to construct the annotation sample and the final human annotations was substantial (Cohen's $\kappa = 0.78$). This result provides evidence that the consensus-based weak-labeling procedure was effective for constructing candidate positive and negative annotation pools while retaining human annotation as the final reference standard.

\subsection{Sources of Annotator Disagreement}
\label{sec:annotator_disagreement}
We qualitatively examined the posts for which the two expert annotators assigned different binary labels. Three main patterns explained most disagreements: ambiguity between questioning and asserting misinformation, reliance on personal experiences or alternative treatments, and uncertainty regarding the evidentiary and cancer-related boundaries of the annotation task. Table~\ref{tab:disagreement_themes} presents these patterns with representative examples.

\begin{table*}[t]
\centering
\small
\renewcommand{\arraystretch}{1.15}
\setlength{\tabcolsep}{4pt}

\begin{tabularx}{\textwidth}{
    >{\raggedright\arraybackslash}p{0.48\textwidth}
    >{\raggedright\arraybackslash}X
}
\toprule
\textbf{Pattern} & \textbf{Representative examples} \\
\midrule

\textbf{Questioning or speculating versus asserting misinformation:} \par\smallskip
Several posts introduced questionable propositions as questions, possibilities, or uncertain beliefs rather than established facts. These examples may be interpreted either as spreading an unsupported proposition or as seeking clarification without endorsing the proposition.
&
1) ``Random idea! Epsom salt soak for inflamed prostate? Anyone tried this for their prostate? \ldots maybe it'd work for the prostate?''
\par\medskip
2) ``Maybe hope to find ways to control my hormones naturally or follow my oncologist's suggestion.'' \\

\addlinespace[5pt]

\textbf{Personal anecdotes and alternative treatments:} \par\smallskip
Several posts combined personal experiences with implied medical conclusions. These posts blur the distinction between reporting an individual experience, discussing complementary practices, and recommending an unsupported medical treatment.
&
1) ``I found a holistic MD that prescribed ivermectin and it had a wonderful side effect of decreasing my prostate enlargement.'' 
\par\medskip
2) ``I've stopped all regular medication except tamoxifen. Now I'm relying on what I've got: acupuncture, homeopathy, movement.'' \\

\addlinespace[5pt]

\textbf{Unclear evidentiary or cancer-related boundaries:} \par\smallskip
Some claims relied on preliminary evidence, lacked sufficient context, or were only indirectly related to cancer. These cases create uncertainty about whether a claim is sufficiently unsupported, misleading, and directly related to cancer to satisfy the annotation criteria.
&
1) ``They exposed breast cancer cells to the blessed seed extracts; the breast cancer cells were inactivated, whereby they concluded that the blessed seed had promising results in the field of prevention and treatment of cancer.''
\par\medskip
2) ``This doctor had prostate problems, and he developed a secret investigation about this, asking a NASA officer how astronauts pee with no gravity. They drink iodine. Wakame, neem. Saw palmetto. Kelp. Shilajit. Bladderwrack. Anyone tried any of them?'' \\

\bottomrule
\end{tabularx}
\caption{Representative patterns of disagreement between expert annotators.}
\label{tab:disagreement_themes}
\end{table*}

\section{Using LLMs for Classification}\label{sec:full_classificaion}

\subsection{Misinformation Classification}\label{sec:misinfo_class_extra}

To identify a robust model configuration for misinformation classification, we evaluated 21 LLMs across zero-shot and few-shot settings. For each model, we varied the number of in-context examples from $k=0$ (zero-shot) to $k=5$, using the same classification instructions and annotation definition across configurations. All experiments were conducted with a temperature of 0 to reduce output variability. Few-shot instances were drawn from a fixed pool of 10 human-annotated posts excluded from all evaluations, resulting in a held-out set of $N=390$. Table~\ref{tab:model_performance} reports performance across models and shot settings.
This comparison allowed us to assess differences across model families and the effect of the number of in-context examples on classification performance. Because misinformation classification requires distinguishing unsupported or misleading medical claims from questions, personal experiences, and legitimate health information, we considered classification performance, specificity, stability across shot settings, and inference cost. Balancing these criteria, we selected GPT-4o with $k=1$ (F1 = 0.891; TPR = 0.98) and Gemini 3.0 Flash with $k=5$ (F1 = 0.893; TPR = 0.98) as the two final classifiers and applied them independently to the full corpus. Although GPT-5.5 achieved the highest weighted F1 (0.896 at $k=2$), it was not selected because of its higher inference cost and generally lower TPR relative to the selected configurations. We subsequently evaluated the individual predictions and cross-model agreement of the two selected classifiers to determine the final labeling strategy, as described in \S~\ref{sec:classifier_evaluation}--\ref{sec:downstream_label_selection}.

\begin{table*}[h!]
\centering
\small
\begin{minipage}{0.45\textwidth}
\centering
\resizebox{\linewidth}{!}{
\begin{tabular}{llccccc}
\toprule
\textbf{Model} & \textbf{K} & \textbf{P} & \textbf{R} & \textbf{F1} & \textbf{TPR} & \textbf{TNR} \\
\midrule
GPT-5.5 & \multirow{21}{*}{0} & 0.89 & 0.88 & 0.885 & 0.91 & 0.87 \\
GPT-5.4 &  & 0.82 & 0.82 & 0.82 & 0.77 & 0.85 \\
GPT-5.4-mini &  & 0.87 & 0.86 & 0.86 & 0.90 & 0.83 \\
GPT-5.4-nano &  & 0.81 & 0.65 & 0.63 & 1.00 & 0.41 \\
GPT-5.2 &  & 0.87 & 0.86 & 0.86 & 0.88 & 0.84 \\
GPT-5.1 &  & 0.84 & 0.84 & 0.84 & 0.83 & 0.84 \\
GPT-5 &  & 0.87 & 0.87 & 0.87 & 0.87 & 0.87 \\
GPT-5-mini &  & 0.88 & 0.85 & 0.86 & 0.96 & 0.79 \\
GPT-5-nano &  & 0.83 & 0.82 & 0.82 & 0.81 & 0.83 \\
GPT-4.1 &  & 0.85 & 0.84 & 0.84 & 0.85 & 0.83 \\
GPT-4.1-mini &  & 0.87 & 0.86 & 0.87 & 0.90 & 0.84 \\
GPT-4.1-nano &  & 0.80 & 0.78 & 0.79 & 0.84 & 0.74 \\
GPT-4o &  & 0.90 & 0.89 & \textbf{0.888} & 0.97 & 0.83 \\
GPT-4o-mini &  & 0.88 & 0.86 & 0.86 & 0.97 & 0.79 \\
Claude-sonnet-4-6 &  & 0.86 & 0.85 & 0.85 & 0.87 & 0.83 \\
Claude-haiku-4-5 &  & 0.85 & 0.85 & 0.85 & 0.87 & 0.84 \\
Gemini 3.0 Flash &  & 0.90 & 0.89 & \textbf{0.888} & 0.97 & 0.83 \\
Gemini 2.5 Flash &  & 0.87 & 0.84 & 0.85 & 0.94 & 0.78 \\
Qwen3.6-27B &  & 0.82 & 0.82 & 0.82 & 0.78 & 0.85 \\
Meta-Llama-3.1-8B &  & 0.78 & 0.70 & 0.70 & 0.92 & 0.56 \\
Gemma-4-31B-it &  & 0.87 & 0.86 & 0.86 & 0.92 & 0.82 \\
\midrule
GPT-5.5 & \multirow{21}{*}{1} & 0.89 & 0.88 & 0.878 & 0.92 & 0.85 \\
GPT-5.4 &  & 0.82 & 0.82 & 0.82 & 0.78 & 0.85 \\
GPT-5.4-mini &  & 0.86 & 0.85 & 0.85 & 0.88 & 0.84 \\
GPT-5.4-nano &  & 0.88 & 0.83 & 0.83 & 0.99 & 0.72 \\
GPT-5.2 &  & 0.89 & 0.88 & 0.88 & 0.94 & 0.84 \\
GPT-5.1 &  & 0.84 & 0.84 & 0.84 & 0.84 & 0.84 \\
GPT-5 &  & 0.88 & 0.87 & 0.87 & 0.89 & 0.86 \\
GPT-5-mini &  & 0.88 & 0.86 & 0.86 & 0.95 & 0.80 \\
GPT-5-nano &  & 0.83 & 0.83 & 0.83 & 0.79 & 0.85 \\
GPT-4.1 &  & 0.88 & 0.87 & 0.87 & 0.94 & 0.82 \\
GPT-4.1-mini &  & 0.86 & 0.85 & 0.85 & 0.88 & 0.84 \\
GPT-4.1-nano &  & 0.80 & 0.75 & 0.75 & 0.90 & 0.64 \\
GPT-4o &  & 0.91 & 0.89 & \textbf{0.891} & \textbf{0.98} & 0.83 \\
GPT-4o-mini &  & 0.90 & 0.88 & 0.88 & 0.99 & 0.80 \\
Claude-sonnet-4-6 &  & 0.88 & 0.86 & 0.86 & 0.96 & 0.79 \\
Claude-haiku-4-5 &  & 0.85 & 0.85 & 0.85 & 0.85 & 0.85 \\
Gemini 3.0 Flash &  & 0.90 & 0.88 & 0.883 & 0.97 & 0.82 \\
Gemini 2.5 Flash &  & 0.87 & 0.85 & 0.85 & 0.94 & 0.79 \\
Qwen3.6-27B &  & 0.88 & 0.83 & 0.83 & 0.99 & 0.73 \\
Meta-Llama-3.1-8B &  & 0.87 & 0.84 & 0.84 & 0.95 & 0.77 \\
Gemma-4-31B-it &  & 0.88 & 0.86 & 0.86 & 0.93 & 0.82 \\
\midrule
GPT-5.5 & \multirow{21}{*}{2} & 0.91 & 0.89 & \textbf{0.896} & 0.96 & 0.85 \\
GPT-5.4 &  & 0.84 & 0.84 & 0.84 & 0.82 & 0.85 \\
GPT-5.4-mini &  & 0.86 & 0.85 & 0.85 & 0.88 & 0.84 \\
GPT-5.4-nano &  & 0.88 & 0.82 & 0.82 & 1.00 & 0.70 \\
GPT-5.2 &  & 0.90 & 0.88 & 0.88 & 0.97 & 0.82 \\
GPT-5.1 &  & 0.88 & 0.87 & 0.87 & 0.93 & 0.83 \\
GPT-5 &  & 0.87 & 0.87 & 0.87 & 0.89 & 0.85 \\
GPT-5-mini &  & 0.88 & 0.86 & 0.86 & 0.95 & 0.80 \\
GPT-5-nano &  & 0.81 & 0.81 & 0.81 & 0.76 & 0.84 \\
GPT-4.1 &  & 0.89 & 0.87 & 0.88 & 0.95 & 0.82 \\
GPT-4.1-mini &  & 0.87 & 0.87 & 0.87 & 0.90 & 0.84 \\
GPT-4.1-nano &  & 0.79 & 0.66 & 0.65 & 0.96 & 0.47 \\
GPT-4o &  & 0.91 & 0.89 & 0.888 & 0.98 & 0.82 \\
GPT-4o-mini &  & 0.90 & 0.87 & 0.87 & 0.99 & 0.79 \\
Claude-sonnet-4-6 &  & 0.89 & 0.88 & 0.88 & 0.96 & 0.82 \\
Claude-haiku-4-5 &  & 0.87 & 0.86 & 0.86 & 0.90 & 0.83 \\
Gemini 3.0 Flash &  & 0.91 & 0.89 & \textbf{0.893} & 0.97 & 0.84 \\
Gemini 2.5 Flash &  & 0.88 & 0.86 & 0.86 & 0.95 & 0.79 \\
Qwen3.6-27B &  & 0.89 & 0.86 & 0.87 & 0.98 & 0.79 \\
Meta-Llama-3.1-8B &  & 0.87 & 0.85 & 0.85 & 0.93 & 0.80 \\
Gemma-4-31B-it &  & 0.88 & 0.87 & 0.87 & 0.94 & 0.82 \\

\bottomrule
\end{tabular}
}
\\
(a) Results for $k=0$--$2$
\end{minipage}
\hfill
\begin{minipage}{0.45\textwidth}
\centering
\resizebox{\linewidth}{!}{
\begin{tabular}{llccccc}
\toprule
\textbf{Model} & \textbf{K} & \textbf{P} & \textbf{R} & \textbf{F1} & \textbf{TPR} & \textbf{TNR} \\
\midrule
GPT-5.5 & \multirow{21}{*}{3} & 0.90 & 0.89 & \textbf{0.893} & 0.95 & 0.85 \\
GPT-5.4 &  & 0.86 & 0.86 & 0.86 & 0.88 & 0.85 \\
GPT-5.4-mini &  & 0.85 & 0.84 & 0.84 & 0.85 & 0.84 \\
GPT-5.4-nano &  & 0.88 & 0.84 & 0.84 & 0.97 & 0.76 \\
GPT-5.2 &  & 0.90 & 0.88 & 0.88 & 0.96 & 0.83 \\
GPT-5.1 &  & 0.87 & 0.86 & 0.86 & 0.92 & 0.82 \\
GPT-5 &  & 0.89 & 0.88 & 0.885 & 0.91 & 0.87 \\
GPT-5-mini &  & 0.88 & 0.86 & 0.86 & 0.96 & 0.79 \\
GPT-5-nano &  & 0.81 & 0.81 & 0.81 & 0.77 & 0.83 \\
GPT-4.1 &  & 0.90 & 0.88 & 0.88 & 0.96 & 0.83 \\
GPT-4.1-mini &  & 0.88 & 0.87 & 0.87 & 0.91 & 0.85 \\
GPT-4.1-nano &  & 0.79 & 0.68 & 0.68 & 0.95 & 0.51 \\
GPT-4o &  & 0.91 & 0.89 & 0.888 & 0.98 & 0.82 \\
GPT-4o-mini &  & 0.90 & 0.88 & 0.88 & 0.99 & 0.81 \\
Claude-sonnet-4-6 &  & 0.88 & 0.86 & 0.87 & 0.94 & 0.82 \\
Claude-haiku-4-5 &  & 0.86 & 0.86 & 0.86 & 0.87 & 0.85 \\
Gemini 3.0 Flash &  & 0.89 & 0.88 & 0.881 & 0.95 & 0.83 \\
Gemini 2.5 Flash &  & 0.88 & 0.86 & 0.86 & 0.96 & 0.79 \\
Qwen3.6-27B &  & 0.89 & 0.85 & 0.86 & 0.98 & 0.77 \\
Meta-Llama-3.1-8B &  & 0.87 & 0.85 & 0.86 & 0.93 & 0.80 \\
Gemma-4-31B-it &  & 0.88 & 0.86 & 0.86 & 0.93 & 0.82 \\
\midrule
GPT-5.5 & \multirow{21}{*}{4} & 0.89 & 0.88 & 0.883 & 0.93 & 0.85 \\
GPT-5.4 &  & 0.87 & 0.86 & 0.87 & 0.91 & 0.83 \\
GPT-5.4-mini &  & 0.85 & 0.84 & 0.84 & 0.87 & 0.83 \\
GPT-5.4-nano &  & 0.88 & 0.84 & 0.84 & 0.99 & 0.73 \\
GPT-5.2 &  & 0.90 & 0.88 & 0.88 & 0.97 & 0.82 \\
GPT-5.1 &  & 0.87 & 0.86 & 0.86 & 0.90 & 0.83 \\
GPT-5 &  & 0.89 & 0.89 & \textbf{0.888} & 0.92 & 0.87 \\
GPT-5-mini &  & 0.89 & 0.87 & 0.87 & 0.96 & 0.81 \\
GPT-5-nano &  & 0.82 & 0.82 & 0.82 & 0.78 & 0.84 \\
GPT-4.1 &  & 0.89 & 0.88 & 0.88 & 0.96 & 0.82 \\
GPT-4.1-mini &  & 0.88 & 0.87 & 0.88 & 0.92 & 0.85 \\
GPT-4.1-nano &  & 0.79 & 0.67 & 0.66 & 0.97 & 0.47 \\
GPT-4o &  & 0.90 & 0.89 & \textbf{0.888} & 0.97 & 0.83 \\
GPT-4o-mini &  & 0.90 & 0.88 & 0.88 & 0.98 & 0.82 \\
Claude-sonnet-4-6 &  & 0.88 & 0.87 & 0.87 & 0.94 & 0.83 \\
Claude-haiku-4-5 &  & 0.87 & 0.86 & 0.86 & 0.88 & 0.84 \\
Gemini 3.0 Flash &  & 0.90 & 0.88 & 0.886 & 0.97 & 0.83 \\
Gemini 2.5 Flash &  & 0.89 & 0.87 & 0.87 & 0.96 & 0.81 \\
Qwen3.6-27B &  & 0.90 & 0.88 & 0.88 & 0.97 & 0.81 \\
Meta-Llama-3.1-8B &  & 0.87 & 0.85 & 0.85 & 0.94 & 0.78 \\
Gemma-4-31B-it &  & 0.88 & 0.87 & 0.87 & 0.94 & 0.82 \\
\midrule
GPT-5.5 & \multirow{21}{*}{5} & 0.90 & 0.89 & \textbf{0.893} & 0.96 & 0.85 \\
GPT-5.4 &  & 0.88 & 0.87 & 0.87 & 0.92 & 0.84 \\
GPT-5.4-mini &  & 0.88 & 0.87 & 0.87 & 0.92 & 0.83 \\
GPT-5.4-nano &  & 0.88 & 0.83 & 0.83 & 0.99 & 0.72 \\
GPT-5.2 &  & 0.89 & 0.88 & 0.88 & 0.94 & 0.84 \\
GPT-5.1 &  & 0.88 & 0.87 & 0.87 & 0.94 & 0.82 \\
GPT-5 &  & 0.88 & 0.88 & 0.88 & 0.91 & 0.85 \\
GPT-5-mini &  & 0.89 & 0.87 & 0.87 & 0.96 & 0.82 \\
GPT-5-nano &  & 0.83 & 0.83 & 0.83 & 0.81 & 0.85 \\
GPT-4.1 &  & 0.89 & 0.88 & 0.88 & 0.95 & 0.84 \\
GPT-4.1-mini &  & 0.87 & 0.87 & 0.87 & 0.90 & 0.85 \\
GPT-4.1-nano &  & 0.79 & 0.67 & 0.66 & 0.96 & 0.48 \\
GPT-4o &  & 0.91 & 0.89 & 0.891 & 0.98 & 0.83 \\
GPT-4o-mini &  & 0.90 & 0.87 & 0.87 & 0.99 & 0.79 \\
Claude-sonnet-4-6 &  & 0.89 & 0.87 & 0.88 & 0.95 & 0.82 \\
Claude-haiku-4-5 &  & 0.87 & 0.87 & 0.87 & 0.90 & 0.85 \\
Gemini 3.0 Flash &  & 0.91 & 0.89 & \textbf{0.893} & \textbf{0.98} & 0.83 \\
Gemini 2.5 Flash &  & 0.89 & 0.87 & 0.87 & 0.97 & 0.80 \\
Qwen3.6-27B &  & 0.90 & 0.88 & 0.88 & 0.97 & 0.82 \\
Meta-Llama-3.1-8B &  & 0.87 & 0.85 & 0.85 & 0.94 & 0.79 \\
Gemma-4-31B-it &  & 0.90 & 0.88 & 0.88 & 0.97 & 0.82 \\
\bottomrule
\end{tabular}
}
\vspace{2mm}
\\
(b) Results for $k=3$--$5$
\end{minipage}

\caption{Performance comparison across models and shot settings. P, R, and F1 show weighted precision, weighted recall, and weighted F1, respectively; TPR denotes the true-positive rate for the misinformation class, and TNR denotes the true-negative rate for the non-misinformation class. We select models from OpenAI (GPT-5.5[April 23, 2026], GPT-5.4[March 5, 2026], GPT-5.4-mini[March 17, 2026], GPT-5.4-nano[March 17, 2026], GPT-5.2[December 11, 2025], GPT-5.1[November 12, 2025], GPT-5[August 7, 2025], GPT-5-mini[August 7, 2025], GPT-5-nano[August 7, 2025], GPT-4.1[April 14, 2025], GPT-4.1-mini[April 14, 2025], GPT-4.1-nano[April 14, 2025], GPT-4o[May 13, 2024], GPT-4o-mini[July 18, 2024]), Claude (Claude-sonnet-4-6[February 17, 2026], Claude-haiku-4-5[October 15, 2025]), Gemini (Gemini 3.0 Flash[December 17, 2025], Gemini 2.5 Flash[April 17, 2025]), Meta-Llama-3.1-8B[July 23, 2024], gemma-4-31B-it[April 2, 2026], and Qwen3.6-27B[April 21, 2026].} 
\label{tab:model_performance}
\end{table*}

\subsubsection{Final Classifier Evaluation}
\label{sec:classifier_evaluation}

We evaluated two final LLM classifiers against the held-out human
annotations: GPT-4o, hereafter Model~1 ($M_1$), and Gemini 3.0 Flash,
hereafter Model~2 ($M_2$). Because the annotation set was constructed
through stratified sampling rather than random sampling from the corpus,
these results characterize classification performance on the evaluation set and should not be interpreted as estimates of corpus-level
misinformation prevalence.
As shown in Table~\ref{tab:model_agreement_eval}, the two classifiers
performed comparably against the human annotations. $M_1$ achieved a weighted
F1 score of 0.891, while $M_2$ achieved 0.893. Both models achieved a TPR
of 0.981, with TNRs of 0.829 and 0.833, respectively. Given their similar performance, neither individual classifier was clearly preferable as the sole labeling model.
We therefore evaluated two model-agreement decision rules
against the same held-out human annotations. Under the
\textit{intersection} rule, a post was assigned a positive misinformation
label only when both models classified it as misinformation:

\begin{equation}
    \hat{y}_{\mathrm{int}}
    =
    \mathbb{I}(M_1 = 1 \land M_2 = 1).
\end{equation}

Under \textit{union} rule, a positive label was assigned when either
classifier predicted misinformation:

\begin{equation}
    \hat{y}_{\mathrm{union}}
    =
    \mathbb{I}(M_1 = 1 \lor M_2 = 1).
\end{equation}

The intersection achieved the highest weighted F1 (0.896) and specificity
(0.842), while maintaining high sensitivity (0.974) (Table~\ref{tab:model_agreement_eval}). In contrast, the
union increased sensitivity to 0.987 but reduced specificity to 0.821 and
produced the lowest weighted F1 (0.888). Although differences in aggregate
performance were small, the intersection provided a more conservative
operating point. We therefore adopted the intersection
rule for constructing the high-confidence misinformation set used in
downstream analyses.

\begin{table}[h]
\centering
\small
\resizebox{\columnwidth}{!}{%
\begin{tabular}{lccccc}
\toprule
\textbf{Strategy} & \textbf{P} & \textbf{R} & \textbf{F1} &
\textbf{TPR} & \textbf{TNR} \\
\midrule
$M_1$
    & .908 & .889 & .891 & .981 & .829 \\ 
$M_2$
    & .910 & .892 & .893 & .981 & .833 \\ 
Intersection ($M_1 \land M_2$)
    & \textbf{.910} & \textbf{.895} & \textbf{.896}
    & .974 & \textbf{.842} \\
Union ($M_1 \lor M_2$)
    & .908 & .887 & .888 & \textbf{.987} & .821 \\
\bottomrule
\end{tabular}%
}
\caption{Performance of the individual classifiers and agreement-based
decision rules against the held-out human-annotated evaluation set
($N=390$). P, R, and F1 denote weighted precision, recall, and F1,
respectively; TPR denotes true-positive rate for misinformation, and TNR denotes the true-negative rate for non-misinformation.}
\label{tab:model_agreement_eval}
\end{table}

\subsubsection{Cross-Model Agreement on the Full Corpus}\label{sec:full_corpus_agreement}

We next examined agreement between the two final classifiers across all \corpusTwo{} posts. Table~\ref{tab:full_corpus_agreement} reports their classification patterns. $M_1$ classified 2,933 posts (2.19\%) as misinformation, whereas $M_2$ classified 3,680 posts (2.75\%). The models jointly classified 1,730 posts (1.29\%) as misinformation and 128,930 posts (96.35\%) as non-misinformation.
Overall, the models disagreed on 3,153 posts (2.36\% of the corpus): 1,203 posts were classified as misinformation by $M_1$ only and 1,950 by $M_2$ only. Thus, corpus-level agreement was high, largely because both models classified the substantial majority of posts as non-misinformation. Agreement on positive classifications was substantially lower. Among the 4,883 posts classified as misinformation by at least one model, 1,730 (35.43\%) were identified by both models, 1,203 (24.64\%) by $M_1$ only, and 1,950 (39.93\%) by $M_2$ only. Thus, high overall agreement does not imply equivalent decision boundaries for the comparatively rare misinformation class.

\begin{table}[t]
\centering
\small
\begin{tabular}{lrr}
\toprule
\textbf{Classification} & \textbf{$N$} & \textbf{\% Corpus} \\
\midrule
$M_1$ positive
    & 2,933   & 2.19 \\
$M_2$ positive
    & 3,680   & 2.75 \\
\midrule
Both positive ($M_1{+},M_2{+}$)
    & 1,730   & 1.29 \\
$M_1$ only ($M_1{+},M_2{-}$)
    & 1,203   & 0.90 \\
$M_2$ only ($M_1{-},M_2{+}$)
    & 1,950   & 1.46 \\
Both negative ($M_1{-},M_2{-}$)
    & 128,930 & 96.35 \\
\midrule
Any disagreement
    & 3,153   & 2.36 \\
\bottomrule
\end{tabular}
\caption{Agreement between the two final classifiers across the full corpus ($N=\corpusTwo$). Individual model-positive counts overlap; the four model-combination rows are mutually exclusive.}
\label{tab:full_corpus_agreement}
\end{table}

\subsubsection{Label Selection for Downstream Analysis}
\label{sec:downstream_label_selection}
For the primary downstream analyses, we adopted the intersection rule and retained the 1,730 posts classified as misinformation by both models. We refer to this subset as \textit{high-confidence misinformation}. This terminology denotes cross-model agreement and does not imply that model agreement constitutes human-verified ground truth. 
Posts for which the models disagree (3,153) are treated as
model-uncertain cases rather than assigned to either high-confidence group. Since our analyses require clearly separated misinformation and non-misinformation groups, these cases were excluded from the primary comparison.
Importantly, the 1.29\% cross-model positive rate should not be interpreted as an estimate of the true prevalence of misinformation in the corpus.

\subsubsection{Methodological Considerations}
\label{sec:annotation_limitations}
The annotation and classification procedure involves two related
limitations. First, the human-annotated evaluation set was intentionally constructed using weak labels and multi-model consensus to ensure sufficient representation of both misinformation and non-misinformation.
It therefore does not constitute a random sample of the full corpus, and performance on this evaluation set may not directly generalize to the full corpus or reflect performance under the corpus's natural class distribution.
Second, consensus-based sampling may preferentially select clearer cases and underrepresent ambiguous posts near the misinformation decision boundary. This limitation is particularly relevant given that the two final classifiers show greater disagreement within the set of posts identified as misinformation by at least one model. Our use of the intersection for primary downstream analyses reduces reliance on model-specific positive predictions but may exclude valid misinformation identified by only one classifier.

\subsection{Taxonomy Dimension Classification}
Table~\ref{tab:model_task_shot_performance} reports performance across taxonomy dimensions and prompting settings. Performance varied by dimension and model, with few-shot prompting providing substantial gains for more nuanced dimensions such as Misinformation Type, while Misinformation Topic and Cancer Stage achieved the strongest overall performance. For downstream analysis, we selected the model and prompting configuration with the highest validation F1 for each taxonomy dimension. Accordingly, we used Gemini 3.0  with $k=3$ for Cancer Stage, GPT-4o with $k=1$ for Information Type, GPT-4o with $k=4$ for Misinformation Type, Gemini 3 Flash with $k=3$ for Risk Level, Gemini 3.0 with $k=2$ for Stance on Misinformation, and GPT-4o with $k=3$ for Misinformation Topic. 

\begin{table*}[h]
\centering
\small
\resizebox{0.9\textwidth}{!}{%
\begin{tabular}{lcccrrrrr}
\toprule
Dimension & Shots & Eval Data Points & Model & Precision & Recall & F1 & TPR & TNR \\
\midrule
\multirow{10}{*}{Cancer Stage} & \multirow{2}{*}{0} & \multirow{10}{*}{149}
& Gemini 3.0 Flash & 0.91 & 0.83 & 0.86 & 0.83 & 0.95 \\
& & & GPT-4o & 0.93 & 0.54 & 0.64 & 0.54 & 0.95 \\
& \multirow{2}{*}{1} &
& Gemini 3.0 Flash & 0.93 & 0.85 & 0.87 & 0.85 & 0.95 \\
& & & GPT-4o & 0.92 & 0.68 & 0.76 & 0.68 & 0.94 \\
& \multirow{2}{*}{2} &
& Gemini 3.0 Flash & 0.92 & 0.85 & 0.88 & 0.86 & 0.95 \\
& & & GPT-4o & 0.89 & 0.67 & 0.74 & 0.67 & 0.92 \\
& \multirow{2}{*}{3} &
& Gemini 3.0 Flash & 0.92 & 0.88 & \textbf{0.90} & 0.88 & 0.95 \\
& & & GPT-4o & 0.89 & 0.72 & 0.78 & 0.72 & 0.92 \\
& \multirow{2}{*}{4} &
& Gemini 3.0 Flash & 0.91 & 0.85 & 0.88 & 0.85 & 0.95 \\
& & & GPT-4o & 0.88 & 0.67 & 0.74 & 0.67 & 0.94 \\
\midrule
\multirow{10}{*}{Information Type} & \multirow{2}{*}{0} & \multirow{10}{*}{149}
& Gemini 3.0 Flash & 0.85 & 0.79 & 0.81 & 0.79 & 0.84 \\
& & & GPT-4o & 0.87 & 0.87 & 0.86 & 0.87 & 0.87 \\
& \multirow{2}{*}{1} &
& Gemini 3.0 Flash & 0.89 & 0.80 & 0.83 & 0.80 & 0.90 \\
& & & GPT-4o & 0.90 & 0.85 & \textbf{0.87} & 0.85 & 0.91 \\
& \multirow{2}{*}{2} &
& Gemini 3.0 Flash & 0.86 & 0.77 & 0.80 & 0.77 & 0.87 \\
& & & GPT-4o & 0.88 & 0.79 & 0.82 & 0.78 & 0.90 \\
& \multirow{2}{*}{3} &
& Gemini 3.0 Flash & 0.88 & 0.83 & 0.84 & 0.83 & 0.87 \\
& & & GPT-4o & 0.86 & 0.81 & 0.83 & 0.81 & 0.87 \\
& \multirow{2}{*}{4} &
& Gemini 3.0 Flash & 0.85 & 0.78 & 0.80 & 0.78 & 0.84 \\
& & & GPT-4o & 0.87 & 0.79 & 0.83 & 0.79 & 0.90 \\
\midrule
\multirow{10}{*}{Misinformation Type} & \multirow{2}{*}{0} & \multirow{10}{*}{137}
& Gemini 3.0 Flash & 0.81 & 0.10 & 0.10 & 0.10 & 0.95 \\
& & & GPT-4o & 0.86 & 0.13 & 0.12 & 0.13 & 0.95 \\
& \multirow{2}{*}{1} &
& Gemini 3.0 Flash & 0.77 & 0.35 & 0.46 & 0.35 & 0.88 \\
& & & GPT-4o & 0.83 & 0.46 & 0.57 & 0.46 & 0.85 \\
& \multirow{2}{*}{2} &
& Gemini 3.0 Flash & 0.77 & 0.19 & 0.25 & 0.19 & 0.93 \\
& & & GPT-4o & 0.81 & 0.40 & 0.51 & 0.40 & 0.81 \\
& \multirow{2}{*}{3} &
& Gemini 3.0 Flash & 0.78 & 0.25 & 0.29 & 0.25 & 0.93 \\
& & & GPT-4o & 0.83 & 0.62 & 0.70 & 0.62 & 0.82 \\
& \multirow{2}{*}{4} &
& Gemini 3.0 Flash & 0.76 & 0.33 & 0.41 & 0.33 & 0.88 \\
& & & GPT-4o & 0.82 & 0.64 & \textbf{0.71} & 0.64 & 0.82 \\
\midrule
\multirow{10}{*}{Risk Level} & \multirow{2}{*}{0} & \multirow{10}{*}{149}
& Gemini 3.0 Flash & 0.53 & 0.14 & 0.10 & 0.14 & 0.83 \\
& & & GPT-4o & 0.65 & 0.16 & 0.12 & 0.16 & 0.78 \\
& \multirow{2}{*}{1} &
& Gemini 3.0 Flash & 0.62 & 0.25 & 0.26 & 0.25 & 0.82 \\
& & & GPT-4o & 0.62 & 0.17 & 0.16 & 0.17 & 0.80 \\
& \multirow{2}{*}{2} &
& Gemini 3.0 Flash & 0.60 & 0.32 & 0.35 & 0.32 & 0.81 \\
& & & GPT-4o & 0.66 & 0.30 & 0.33 & 0.30 & 0.83 \\
& \multirow{2}{*}{3} &
& Gemini 3.0 Flash & 0.65 & 0.39 & \textbf{0.40} & 0.39 & 0.80 \\
& & & GPT-4o & 0.69 & 0.35 & 0.37 & 0.35 & 0.82 \\
& \multirow{2}{*}{4} &
& Gemini 3.0 Flash & 0.61 & 0.34 & 0.37 & 0.34 & 0.80 \\
& & & GPT-4o & 0.65 & 0.28 & 0.30 & 0.28 & 0.82 \\
\midrule
\multirow{10}{*}{Stance on Misinformation} & \multirow{2}{*}{0} & \multirow{10}{*}{142}
& Gemini 3.0 Flash & 0.71 & 0.60 & 0.62 & 0.60 & 0.87 \\
& & & GPT-4o & 0.59 & 0.40 & 0.40 & 0.40 & 0.88 \\
& \multirow{2}{*}{1} &
& Gemini 3.0 Flash & 0.66 & 0.61 & 0.61 & 0.61 & 0.83 \\
& & & GPT-4o & 0.76 & 0.52 & 0.57 & 0.52 & 0.91 \\
& \multirow{2}{*}{2} &
& Gemini 3.0 Flash & 0.77 & 0.65 & \textbf{0.67} & 0.65 & 0.90 \\
& & & GPT-4o & 0.69 & 0.45 & 0.45 & 0.45 & 0.90 \\
& \multirow{2}{*}{3} &
& Gemini 3.0 Flash & 0.73 & 0.63 & 0.64 & 0.63 & 0.88 \\
& & & GPT-4o & 0.72 & 0.52 & 0.55 & 0.52 & 0.90 \\
& \multirow{2}{*}{4} &
& Gemini 3.0 Flash & 0.72 & 0.72 & 0.63 & 0.62 & 0.88 \\
& & & GPT-4o & 0.72 & 0.49 & 0.52 & 0.49 & 0.90 \\
\midrule
\multirow{10}{*}{Misinformation Topic} & \multirow{2}{*}{0} & \multirow{10}{*}{130}
& Gemini 3.0 Flash & 0.86 & 0.85 & 0.84 & 0.85 & 0.84 \\
& & & GPT-4o & 0.96 & 0.95 & 0.95 & 0.95 & 0.92 \\
& \multirow{2}{*}{1} &
& Gemini 3.0 Flash & 0.86 & 0.85 & 0.84 & 0.85 & 0.81 \\
& & & GPT-4o & 0.94 & 0.94 & 0.94 & 0.94 & 0.84 \\
& \multirow{2}{*}{2} &
& Gemini 3.0 Flash & 0.85 & 0.83 & 0.82 & 0.83 & 0.79 \\
& & & GPT-4o & 0.95 & 0.94 & 0.95 & 0.94 & 0.88 \\
& \multirow{2}{*}{3} &
& Gemini 3.0 Flash & 0.85 & 0.82 & 0.81 & 0.82 & 0.81 \\
& & & GPT-4o & 0.96 & 0.96 & \textbf{0.96} & 0.96 & 0.91 \\
& \multirow{2}{*}{4} &
& Gemini 3.0 Flash & 0.88 & 0.83 & 0.85 & 0.83 & 0.86 \\
& & & GPT-4o & 0.95 & 0.94 & 0.95 & 0.94 & 0.88 \\
\bottomrule
\end{tabular}%
}
\caption{Model performance across dimensions \& shot settings. We select models from OpenAI (GPT-4o [August 7, 2025]) and Gemini (Gemini 3.0 Flash [December 17, 2025]). For TPR and TNR, weighted scores are reported.}
\label{tab:model_task_shot_performance}
\end{table*}

\section{Prompts}
\label{sec:taxonomy_prompts}

The complete prompts used for the seven taxonomy dimensions are provided
below.

\begin{promptbox}{Dimension 1: Presence of Misinformation}
You are a medical expert, specialized in identifying cancer-related misinformation.

Definition:

Cancer misinformation refers to any information about cancer, its risk factors, prevention, diagnosis, or treatment that is false, inaccurate, misleading, or not supported by current scientific evidence or consensus, and that may lead to harmful consequences. This includes:

\textit{Unproven or ineffective treatments} \\
\textit{Misrepresentation or selective use of facts} \\
\textit{Outdated or decontextualized medical information} \\
\textit{Conspiracy narratives (e.g., hidden cures)} \\
\textit{Promotional or commercial content presented as medical advice}

Task: Analyze the text provided below. Classify it as either:

`Yes' if the text contains misinformation

`No' if the text does not contain misinformation or if there is insufficient information

Output Format:
Label: `Yes' or `No'. Don't use additional text, and don't use [] around the label \\

Evidence: If classified as 1, quote the specific part of the text that suggests or implies misinformation and explain why it contradicts scientific evidence. \\

If classified as 1, explain your step-by-step reasoning to conclude that the text contains misinformation
\end{promptbox}

\begin{promptbox}{Dimension 2: Cancer Stage}
You are an expert medical data annotator.

Your task is to classify the text into the most relevant stage of the cancer care journey.

Here are the categories:

- \textit{Prevention/Screening}:
Risk factors, lifestyle, prevention behaviors, early detection, screening, mammograms, colonoscopies, genetic risk, or preventive advice.

- \textit{Treatment}:
Medical interventions such as chemotherapy, radiation, surgery, immunotherapy, medications, or active treatment decisions.

- \textit{Other/Unclear/NA}:
Diagnosis, survivorship, post-treatment experiences, emotional support, unclear references, unrelated discussion, or insufficient information. 

Instructions: \\
- Choose the single BEST category. \\
- Focus on the primary topic of the text. \\
- If multiple stages are mentioned, prioritize the main emphasis. \\

Output format:
`Prevention/Screening'
OR
`Treatment'
OR
`Other/Unclear/NA'

No additional words.
\end{promptbox}

\begin{promptbox}{Dimension 3: Information Type}
You are an expert medical data annotator.

Your task is to classify the communication style of the text regarding cancer misinformation.

Definitions:

- \textit{Seeking}:
The user is asking for advice, clarification, information, recommendations, or a second opinion.

- \textit{Sharing}:
The user is providing information, opinions, claims, experiences, or statements.

- \textit{Both}:
The text simultaneously includes both seeking and sharing behavior.

Instructions: \\
- Focus on the author's communicative intent. \\
- Questions alone are not always ``Seeking'' if the text mainly shares information. \\
- If the text strongly contains both behaviors, choose ``Both''.

Output format:
Seeking
OR
Sharing
OR
Both

No additional words.
\end{promptbox}

\begin{promptbox}{Dimension 4: Misinformation Type}
You are an expert medical data annotator.

Your task is to classify the PRIMARY type of cancer-related misinformation present in the text.

Here are the definitions:

- \textit{Fabricated Content}:
Completely false information with no factual basis.

- \textit{Misleading Content}:
Information that distorts, exaggerates, oversimplifies, or selectively presents facts.

- \textit{False Context}:
True information shared in an incorrect, misleading, or inappropriate context.

- \textit{False Connection}:
The headline, claim, or conclusion does not match the supporting content.

- \textit{Imposter Content}:
Fake, impersonated, or falsely attributed credible sources.

- \textit{Manipulated Content}:
Altered images, videos, audio, statistics, or data.

- \textit{Satire/Parody}:
Content intended humorously or satirically that could be misunderstood as factual.

- \textit{}Other:
Does not clearly fit the categories above. 

Instructions: \\ 
- Select the dominant misinformation type. \\
- Focus on HOW the misinformation is constructed, not the topic itself. \\
- If no clear misinformation type is identifiable, select ``Other''.

Output format:
[Label]

No additional words.
\end{promptbox}

\begin{promptbox}{Dimension 5: Risk Level}
You are an expert medical data annotator.

Your task is to classify the potential health risk level of the misinformation in the text.

Definitions:

\textit{Low Risk}:
Content with little or no likely impact on health behavior or outcomes; trivial, vague, or unlikely to influence medical decisions.

\textit{Moderate Risk}:
Content that may influence beliefs, perceptions, or minor health-related decisions.

\textit{High Risk}:
Content likely to encourage harmful health decisions, treatment avoidance, delayed care, dangerous alternative treatments, or potentially life-threatening behavior.

Instructions: \\
- Evaluate the likely real-world consequences if someone believed or acted on the misinformation. \\
- Consider severity, immediacy, and potential harm. \\
- Focus on potential behavioral impact, not offensiveness. \\

Output format:
`Low'
OR
`Moderate'
OR
`High'

No additional words.
\end{promptbox}

\begin{promptbox}{Dimension 6: Stance on Misinformation}
You are an expert medical data annotator.

Your task is to classify the stance expressed toward cancer-related misinformation.

Here are the definitions: 

\textit{Accepts}:
The text explicitly endorses, believes, supports, or promotes the misinformation claim.

\textit{Accept + Question}:
The text leans toward accepting or considering the misinformation while expressing minor hesitation or asking clarifying questions.

\textit{Rejects}:
The text explicitly refutes, debunks, criticizes, or disagrees with the misinformation claim.

\textit{Questions}:
The text expresses curiosity, uncertainty, skepticism, or asks questions without clearly accepting or rejecting the claim.

\textit{Reject + Question}:
The text fundamentally rejects the misinformation but also asks a related follow-up question.

Instructions: \\ 
- Focus on the author's stance toward the misinformation itself. \\
- Distinguish genuine curiosity from endorsement. \\
- Choose the closest overall stance. \\

Output format:
`Accepts'
OR
`Accept + Question'
OR
`Rejects'
OR
`Questions'
OR
`Reject + Question'

No additional words.
\end{promptbox}

\begin{promptbox}{Dimension 7: Misinformation Topic}
You are an expert medical data annotator.

Your task is to classify the PRIMARY topic of cancer-related misinformation discussed in the text.

Here are the categories:

\textit{Causation, risk factors \& prevention claims}:
(Unproven causes of cancer, misrepresented risk factors, unsupported prevention methods, e.g., ``sugar causes cancer,'' ``alkaline diet prevents cancer,'' ``stress causes tumors.'') \\

\textit{Diagnosis \& screening claims}:
(Claims about symptoms, self-diagnosis, screening tests being dangerous, inaccurate, unnecessary, or misleading.) 

\textit{Safety \& effectiveness of conventional treatments}:
(Claims that chemotherapy, radiation, surgery, or medical treatments are ineffective, harmful, toxic, unnecessary, or intentionally deceptive.) 

\textit{Unproven or alternative treatments}:
(Herbs, supplements, diets, naturopathy, homeopathy, detoxes, ``natural cures,'' or non-evidence-based alternatives presented as cancer treatments.) 

\textit{Conspiracy \& institutional distrust}:
(Big Pharma conspiracies, hidden cures, government cover-ups, suppressed research, distrust of doctors or institutions.) 

\textit{Morality, religion \& ideology}:
(Religious, spiritual, karmic, moral, or ideological explanations or objections related to cancer and treatment.) 

\textit{Medical autonomy \& medical freedom}:
(Rights-based arguments about refusing treatment, parental authority, bodily autonomy, or distrust of medical mandates.) 

Instructions:
- Select the SINGLE dominant topic. \\
- Focus on the central misinformation theme. \\
- If multiple topics appear, choose the primary emphasis. \\
\\
Output format:
[Label]

No additional words.
\end{promptbox}

\begin{center}
\captionof{figure}{Prompts used for the seven taxonomy dimensions.}
\label{fig:taxonomy_prompts}
\end{center}


\section{Label Distribution}
\subsection{Ground-Truth Label Distribution}

Table~\ref{tab:annotation_distribution} presents the distribution of labels across the seven taxonomy dimensions in the human-annotated ground-truth dataset. Misinformation Presence is reported across all 400 annotated posts, whereas the remaining dimensions are reported for the 161 posts with a final misinformation label.

\begin{table*}[h]
\centering
\small
\renewcommand{\arraystretch}{1.3}

\resizebox{0.9\textwidth}{!}{%
\begin{tabular}{p{0.25\textwidth}p{0.45\textwidth}rr}

\toprule
\textbf{Dimension} & \textbf{Class} & \textbf{Raw Number} & \textbf{Percentage} \\
\midrule

\multirow{3}{0.25\textwidth}{Misinformation Presence}
& Misinformation & 161 & 40.25\% \\
& Not Misinformation & 235 & 58.75\% \\
& Unclear & 4 & 1.00\% \\

\midrule

\multirow{3}{0.25\textwidth}{Cancer Stage}
& Treatment & 113 & 70\% \\
& Prevention / Screening & 35 & 22\% \\
& Other / Unclear / NA & 13 & 8\% \\

\midrule

\multirow{3}{0.25\textwidth}{Information Type}
& Both & 83 & 52\% \\
& Sharing & 70 & 43\% \\
& Seeking & 8 & 5\% \\

\midrule

\multirow{5}{0.25\textwidth}{Misinformation Type}
& Fabricated Content & 116 & 72\% \\
& Imposter Content & 16 & 10\% \\
& Other & 13 & 8\% \\
& Misleading Content & 12 & 7\% \\
& False Context & 8 & 5\% \\

\midrule

\multirow{3}{0.25\textwidth}{Risk Level}
& Low Risk & 93 & 58\% \\
& Moderate Risk & 50 & 31\% \\
& High Risk & 18 & 11\% \\

\midrule

\multirow{6}{0.25\textwidth}{Stance on Misinformation}
& Accept + Question & 70 & 43\% \\
& Accept & 60 & 37\% \\
& Question & 17 & 11\% \\
& Reject & 11 & 7\% \\
& Other / Unclear & 2 & 1\% \\
& Reject + Question & 1 & 1\% \\

\midrule

\multirow{7}{0.25\textwidth}{Misinformation Topic}
& Unproven or Alternative Treatments & 100 & 62\% \\
& Causation, Risk Factors, and Prevention Claims & 32 & 20\% \\
& Safety and Effectiveness of Conventional Treatments & 21 & 13\% \\
& Diagnosis and Screening Claims & 10 & 6\% \\
& Conspiracy and Institutional Distrust & 4 & 2\% \\
& Other & 3 & 2\% \\
& Medical Autonomy and Medical Freedom & 1 & 1\% \\

\bottomrule
\end{tabular}%
}

\caption{Distribution of labels across the seven taxonomy dimensions in the ground-truth data. Misinformation Presence is reported over all annotated posts ($N=400$); the remaining dimensions are reported over posts with a final misinformation label ($N=161$). Percentages are calculated using the corresponding denominator for each dimension. Misinformation Type and Misinformation Topic permit multiple labels per post; therefore, their counts and percentages may sum to more than $N$ and 100\%, respectively.}
\label{tab:annotation_distribution}
\end{table*}

\subsection{Courpus Label Distribution}

Table~\ref{tab:taxonomy_distribution} presents the label distributions across taxonomy dimensions for the \corpusmisinfo{} high-confidence misinformation posts.
\begin{table*}[h]
\centering
\resizebox{0.85\textwidth}{!}{
\renewcommand{\arraystretch}{1.2}
\begin{tabular}{p{3.9cm}p{7.5cm}cc}
\hline
\textbf{Dimensions} & \textbf{Label} & \textbf{Count} & \textbf{Percentage} \\
\hline

\multirow{3}{*}{Cancer Stage}
& Treatment & 827 & 47.80 \\
& Prevention/Screening & 334 & 19.31 \\
& Other/Unclear/NA & 569 & 32.89 \\
\hline

\multirow{3}{*}{Information Type}
& Both & 845 & 48.84 \\
& Seeking & 365 & 21.10 \\
& Sharing & 520 & 30.06 \\
\hline

\multirow{8}{*}{Misinformation Type}
& Fabricated Content & 508 & 29.36 \\
& Misleading Content & 175 & 10.12 \\
& False Context & 512 & 29.60 \\
& False Connection & 0 & 0.00 \\
& Imposter Content & 105 & 6.00 \\
& Manipulated Content & 0 & 0.00 \\
& Satire/Parody & 6 & 0.35 \\
& Other & 474 & 27.40 \\
\hline

\multirow{3}{*}{Risk Level}
& High & 274 & 15.84 \\
& Moderate & 940 & 54.34 \\
& Low & 516 & 29.83 \\
\hline

\multirow{6}{*}{Stance on Misinformation}
& Question & 556 & 32.14 \\
& Accept & 510 & 29.48 \\
& Accept + Question & 353 & 20.40 \\
& Reject & 114 & 6.59 \\
& Other / Unclear & 157 & 9.08 \\
& Reject + Question & 40 & 2.31 \\
\hline

\multirow{7}{*}{Misinformation Topic}
& Diagnosis \& screening claims & 292 & 16.88 \\
& Safety \& effectiveness of conventional treatments & 268 & 15.50 \\
& Unproven or alternative treatments & 778 & 44.98 \\
& Causation, risk factors \& prevention claims & 122 & 7.06 \\
& Conspiracy \& institutional distrust & 79 & 4.57 \\
& Medical autonomy \& medical freedom & 38 & 2.20 \\
& Morality, religion \& ideology & 22 & 1.27 \\
& Other & 160 & 9.25 \\
\bottomrule
\end{tabular}}
\caption{Distribution of labels across taxonomy dimensions in the 
LLM-annotated misinformation dataset. The model chose multiple labels for some posts when labeling Misinformation Type and Misinformation Topic; therefore, category counts and percentages for these dimensions may sum to more than $N$ and 100\%, respectively.}
\label{tab:taxonomy_distribution}
\end{table*}

\section{Characterizing Cancer Misinformation}
\label{sec:misinformation_characterization}

\subsection{Subreddit-Level Distribution of Misinformation} \label{sec:subreddit_distribution}

Table~\ref{tab:subreddit_distribution} presents the distribution of high-confidence misinformation across the selected cancer-related subreddits. For each subreddit, we report the number of posts classified as misinformation and its proportion relative to the total number of posts collected from that subreddit, allowing comparison across communities of different sizes.

\begin{table*}[t]
\centering
\scriptsize
\resizebox{0.6\textwidth}{!}{
\begin{tabular}{lrrr}
\toprule
\textbf{Subreddit} & \textbf{Misinfo} & \textbf{Total} & \textbf{Percentage} \\
\midrule
breastcancer            & 667 & 49709 & 1.34 \\
Prostatitis             & 403  & 21150 & 1.91 \\
colonoscopy             & 78  & 17182 & 0.45 \\
doihavebreastcancer     & 107  & 13684 & 0.78 \\
ProstateCancer          & 269 & 13542 & 1.99 \\
coloncancer             & 160  & 9783  & 1.64 \\
lungcancer              & 88  & 4032  & 2.18 \\
LivingWithMBC           & 60  & 3673  & 1.63 \\
BreastCancerSurvivors   & 10   & 539   & 1.86 \\
colorectalcancer        & 5   & 257   & 1.95 \\
LungCancerSupport       & 6    & 245   & 2.45 \\
BreastCancerResearch    & 11   & 53    & 20.75 \\
ProstateCancer\_AS      & 17   & 37    & 45.95 \\
Prostate\_Cancer        & 7    & 14    & 50.00 \\
nsclc                   & 1    & 118   & 0.85 \\
SmallCellLungCancer     & 1    & 63    & 1.59 \\
mtf\_breast\_cancer     & 1    & 27    & 3.70 \\
ProstateTreatment       & 1    & 22    & 4.55 \\
QueerBreastCancer       & 0    & 21    & 0.000 \\
askcoloncancerdoctors   & 0    & 14    & 0.000 \\
ProstateBob             & 0    & 41    & 0.000 \\
BreastCancerGenomics    & 0    & 10    & 0.000 \\
ProstateCancerDesign    & 0    & 2     & 0.000 \\
prostateinfo            & 0    & 1     & 0.000 \\
\bottomrule
\end{tabular}}
\caption{Distribution of high-confidence misinformation across cancer-related subreddits. The misinformation ratio represents the percentage of posts within each subreddit classified as misinformation in our dataset. Ratios for smaller subreddits should be interpreted cautiously due to limited sample sizes.}
\label{tab:subreddit_distribution}
\end{table*}

\subsection{Evidence Theme Analysis}
Table~\ref{tab:reasoning_themes} presents the themes identified from the parts of the Reddit posts that LLMs selected as evidence of cancer-related misinformation. For each theme, the table provides its name and the corresponding description or inclusion criterion.

\begin{table*}[t]
\centering
\small
\resizebox{\textwidth}{!}{%
\renewcommand{\arraystretch}{1.8}
\begin{tabular}{p{3.8cm} p{4.2cm} p{7cm}}
\toprule
\textbf{Themes} & \textbf{Description/Criteria} & \textbf{Example} \\
\midrule

Alternative \& Natural Treatments &
Does the text promote alternative or natural treatments for cancer over conventional methods?  & ``Brew up a cup of green tea, pour into a spray bottle, spray on yourself prior, and let air dry. And have some manuka honey.'', ``Women use Greek yogurt for UTIs and other issues that seem to work wonders for them.'' \\

Cancer Cause Misconceptions &
Does the text provide misleading information about the causes or risk factors of cancer? & ``I had to go back on my prolactin medication because it's the only thing stopping my cancer from growing, but it is also the cause of my cancer.'' \\

Cancer Cure Conspiracies &
Does the text suggest that effective cancer cures are being suppressed or involve conspiracy theories? &  ``chemotherapy can speed cancer spread'', ``I've been reading people online who posted about how chemotherapy is a scam and kills you faster than the cancer.'' \\

Cancer Detection Misconceptions &
Does the text involve misconceptions about cancer detection methods? & ``I have reason to believe that outdated approaches, e.g., premature medical interventions or drugs which impact important hormonal function could cause far more harm than good in some people.'' \\

Cancer Recurrence Concerns &
Does the text discuss concerns about cancer recurrence despite treatment?  &  ``I have this feeling we're going to find out that it wasn't cancer at all that they pulled out.'' \\

Cancer Severity Misunderstandings &
Does the text convey misconceptions about the severity or incurability of certain cancer stages?  & ``She decided to trust ChatGPT to diagnose her skin lesion as 'just a cyst' and canceled her doctor's appointment to check it.'' \\

Cancer Spread Myths &
Does the text mention myths or misconceptions about how cancer spreads? & ``Believe it or not, blood in stool is common, and its etiology is benign 99\% of the time'' and ``Colon cancer is rising, but it's still very rare before 40 years old.''\\

Cancer Statistics Misuse &
Does the text misuse or misrepresent cancer statistics, leading to misconceptions? & ``AI is telling me 4.7\% of mortality over 5yrs.. so does that mean 2.82 months remaining?'' \\

Cancer Treatment Myths &
Does the text contain misleading claims or myths about curing or treating cancer? & ``I have a strong urge that poking anything potentially cancerous in the body breaks its protective shell and could release toxins into the blood.'' , ``I have taken a different path and chosen to delay conventional treatments because when I take them, my body reacts very poorly.'' \\

Dietary Claims &
Does the text claim that specific diets or foods can prevent or cure cancer?  & ``Some scientific research has shown that breast milk kills cancer.'', ``alkaline treatment+baking soda+molasses+water: This suggests the use of this formula as a potential remedy or treatment.'' \\

Emotional \& Psychological Factors &
Does the text claim that emotional or psychological factors are linked to cancer development?  & ``I just looked the percent of early-stage breast cancer eventually metastasizing, and it was 30\%. I'm terrified. This feels like I have a 30\% of surviving this now.'', '"I wish I had never been suckered into treatment and surgery I did not want and should never have had. I should be dead.''\\

Medical Treatment Skepticism &
Does the text express skepticism about conventional cancer treatments or distrust in medical professionals? & ``I saw in an article where it says the radiation can expose one to cancer later in life.'' \\

Cancer-Fighting Benefits &
Does the text assert cancer-fighting benefits that are not backed by evidence? & ``After 3 months of natural treatment, I had my PSA rechecked, and it dropped by 3 points.'' \\

Vaccine and Cancer Myths &
Does the text contain misinformation about vaccines being linked to cancer risk? & ``My husband has been driving me nuts over these stupid claims about covid and vaccine causing cancer.'', ``No doubt a genetic factor that took an external trigger (the vaccine) to start all my testicular issues, despite no issues/pain before that.'' \\

\bottomrule
\end{tabular}}
\caption{Themes identified by LLooM from textual evidence supporting misinformation classifications, with corresponding inclusion criteria and representative evidence examples. Themes characterize the content of the evidence and do not indicate the author's stance toward the claim.}
\label{tab:reasoning_themes}
\end{table*}

\section{Cancer-Related Misinformation Examples}\label{sec:examples}
Table~\ref{tab:misinformation_examples} presents representative Reddit posts containing cancer-related misinformation, while Table~\ref{tab:task_examples} provides examples across the different taxonomy dimensions and labels. The examples include general cancer discussions without reference to a specific cancer type, as well as posts related to breast, lung, colon, and prostate cancer. All examples were paraphrased and anonymized for ethical reporting purposes.

\begin{table*}[h]
\centering
\small
\renewcommand{\arraystretch}{1.15}
\setlength{\tabcolsep}{6pt}

\begin{tabular}{
    >{\raggedright\arraybackslash}p{0.14\textwidth}
    >{\raggedright\arraybackslash}p{0.80\textwidth}
}
\toprule
\textbf{Cancer Type} & \textbf{Example Post} \\
\midrule

General &
*Anyone heard of this thing?* Wild tawtnuk apan mushroom. Not sure how many of you did alternative treatment with standard treatment, but I'm curious about ivermectin, fenben, and that sort of stuff.I was just browsing through Insta, and this thing popped up and got me thinking. Can this be for real? Is it legit? Does it really kill cancer? \\
\addlinespace[3pt]

General &
*Fasting Mimic Diet* Has anyone had experience with the Fasting Mimic Diet during chemo? I’ve heard there is proof this could help shrink tumor size. \\
\addlinespace[3pt]

Breast Cancer &
*Breast cancer, COVID vaccine* Does anyone think getting the COVID vaccine opened the door wide open to their cancer? My mammograms were free and clear until one year following the COVID vaccine and boosters. I can’t prove this, but my slow-growing cancer went from NOTHING to two tumors on the next scan (a year later) that measured 2.5cm. My life completely changed, and I feel like the government used me as a science experiment. \\
\addlinespace[3pt]

Lung Cancer &
*My father has been diagnosed with Stage 3B Lung Cancer.* We are incredibly stressed and desperate for a safe treatment. A relative suggested visiting someone who claims to have cured many cancer patients. The treatment involves a strict 3-month diet of Jowari roti, garlic, and specific spices, along with various Ayurvedic tablets and tonics. We met many patients there who swear they were cured, even those who arrived in very bad condition. We want to believe in this natural path, especially since my father is so weak, but we are scared. Help... \\
\addlinespace[3pt]

Colon Cancer &
*Did eating dark chocolate give me colon cancer?* I'm 29-male no history of cancer in my family, and I got stage 3b colon cancer and feel like it will take me. Every day since 21 I ate dark chocolate; some studies show it is high in lead and metals, and there was even a lawsuit against some manufacturers. Be careful, everyone, but I was very healthy, and I think dark chocolate killed me.\\
\addlinespace[3pt]

Prostate Cancer &
* Off-label treatments* Anyone heard/tried a strict keto or keto carnivore diet with off-label use of fenbendazole to treat recurrence of prostate cancer? \\
\bottomrule
\end{tabular}

\caption{Example Reddit posts containing cancer-related misinformation. Examples are paraphrased and anonymized for ethical reporting purposes.}
\label{tab:misinformation_examples}
\vspace{-0.5cm}
\end{table*}

\begin{table*}[t]
\centering
\scriptsize
\renewcommand{\arraystretch}{1.15}
\setlength{\tabcolsep}{5pt}

\resizebox{\textwidth}{!}{%
\begin{tabular}{
    >{\raggedright\arraybackslash}p{0.12\textwidth}
    >{\raggedright\arraybackslash}p{0.12\textwidth}
    >{\raggedright\arraybackslash}p{0.71\textwidth}
}
\toprule
\textbf{Dimension} & \textbf{Label} & \textbf{Example Text} \\
\midrule

\multirow{2}{=}{\textbf{Cancer Stage}}
& Treatment
& **Father diagnosed with Stage IV colon cancer.** I am shaking as I'm writing this. Yesterday, my father was diagnosed with stage IV colon cancer that has metastasized to his liver. The oncologists basically said there is no cure, but chemotherapy can be administered for palliative care. I refuse to just give up. He has lost a substantial amount of weight over the past year, so he is very weak and thin. I've started doing some research and I've read a few promising testimonials on the effects and benefits of giving him raw carrot juice daily,  approximately 64 ounces a day. Any advice is appreciated. \\
\cmidrule(l){2-3}
& Prevention/Screening
& **When is someone going to address the correlation between the steady use of hair dye and breast cancer.** ALL the women I know who have breast cancer and or succumbed to breast cancer were avid users of hair dye.\\

\addlinespace[5pt]

\multirow{2}{=}{\textbf{Information Type}}
& Seeking
& **Ivermectin / fenbenzanole side effects** I am diagnosed with Gleason 7 4+3 pc with 1 cm sq lesion. Waiting for PSMA, and I’m estimating 2 or 3 months before the treatment actually occurs. In the meantime, I want to try ivermectin / fenbenzanole. I’m NOT going to forgo conventional treatment but use them to perhaps jumpstart my own immune system to deal with the pc. Any suggestions, advice, or warnings, including possible side effects I should be aware of? Shouldn’t make the situation worse, so why not? \\
\cmidrule(l){2-3}
& Sharing
& **Cannabis use during chemo** I am 3/4 treatments into AC, and my only side effect has been hair loss and pain in my fingernails. I have mild bone pain after chemo, but it goes away after a couple of days. I have been an everyday smoker of weed since I was 19. I have continued to use weed through my treatment, and I think it is saving my life. I have no mouth sores. I can still taste food. I still have my appetite. My side effects have been so minimal that it’s hard not to attribute this to the weed.\\

\addlinespace[5pt]

\multirow{3}{=}{\textbf{Misinformation Type}}
& Misleading Content
& **I wonder if all of you who got breast cancer at an early age had any family history or did smoking/drinking?** A lot of doctors say that if you have no family history and you don't smoke/drink, then getting breast cancer is almost impossible. I wonder if it is actually true, or if we are just putting the lives of our young girls at risk out of a false notion? \\
\cmidrule(l){2-3}
& Imposter Content
& **5 Foods To Help With Breast Cancer** Breast cancer is one of the leading causes of death among women. Well, let's consider a few foods that can help women with cancer. Eggs are one of the most potent sources of an essential nutrient known as choline. Not only does it act as brain food, but it may also help lower your risk of breast cancer.\\
\cmidrule(l){2-3}
& Fabricated Content
& **Phones vs cancer** I'm curious to know if anyone developed breast cancer without having positives for hormone testing or genetics… I feel like I got cancer from my iPhone. I always used to put it in my bra on the left side…exactly where my tumor is. I’d only do that when I didn’t have pockets to put my phone in. \\

\addlinespace[5pt]

\multirow{3}{=}{\textbf{Risk Level}}
& High
& **Any advice pls** Doctors recently discovered a quail egg-sized lump in my left breast. I’m terrified. There could be a chance it’s not cancerous, but I’m also scared of even getting a biopsy done. I’ve heard all these theories about cancer metastasis from meddling with the tumor in the first place. Not sure if anyone here has ever tried a holistic approach and actually cured a tumor, but I hear castor oil is great. I don’t even want to have to get a biopsy. I’m really skeptical of any medical providers in this country and don’t trust doctors.\\
\cmidrule(l){2-3}
& Moderate
& *Biopsy or lumpectomy?** I recently had a mass looked at with a mammogram and ultrasound, and it shows I have a 57mm mass that is suspicious for malignancy. They have scheduled me for a biopsy on Thursday. A friend told me they should do a lumpectomy because if it is cancerous and they do a biopsy, it could make it spread and metastasize. So now I’m terrified, and I don’t know what to do. \\
\cmidrule(l){2-3}
& Low
& **5 Foods To Help With Breast Cancer** Breast cancer is one of the leading causes of death among women. Well, let's consider a few foods that can help women with cancer. Eggs are one of the most potent sources of an essential nutrient known as choline. Not only does it act as brain food, but it may also help lower your risk of breast cancer.\\

\addlinespace[5pt]

\multirow{4}{=}{\textbf{Stance on Misinformation}}
& Accepts
& **Fasting shrunk my Prostate**    I went on a seven-day fast, only electrolyte water, and 30-60g collagen protein on the last two days. My prostate shrunk 50\%. When you fast that long, your body is so desperate for food that it starts eating the junk your body can do without, including cancer. \\
\cmidrule(l){2-3}
& Rejects
& **Crazy unsolicited theories about cancer** My dentist told me that maybe I got my cancer because of the CERN and the particle accelerator, and it would also explain the rise in cancer amongst young people. It was such a crazy thing to say, it was almost comical (especially coming from a doctor!!) \\
\cmidrule(l){2-3}
& Accept + Question
& **Is Chrisbeats cancer real?** I saw this website about a guy who beat stage 3 colon cancer naturally. I'm young as well and was thinking about looking into it and seeing if I could beat this naturally as well. I think chemo could do more harm than good, according to his website. Is it legit?\\
\cmidrule(l){2-3}
& Reject + Question
& **Pneumonia recovery / Rife machine use by a lung cancer patient** My dad, who has stage IV NSCLC, recently used a Rife machine for hours and developed a fever, later being hospitalized with severe pneumonia. Does anyone have experience with Rife machine side effects or the recovery process for a lung cancer patient with pneumonia?\\

\addlinespace[5pt]

\multirow{4}{=}{\textbf{Misinformation Topic}}
& Unproven or alternative treatments
& **Homeopathy??** Has anyone used homeopathic methods to assist or replace traditional cancer treatments? A practitioner recommended a strict keto diet, expensive supplements, weekly vitamin C IVs, and hyperbaric oxygen sessions to slow cell multiplication. I feel out of choices—has anyone tried this, and is it actually effective or a waste of money? \\
\cmidrule(l){2-3}
& Diagnosis \& screening claims
& **Denied the bilateral mammogram and opted for just the ultrasound on one breast** I’m not dumb—we all know mammograms don’t show what I need to find. Why put myself through pain and extra radiation? Why aren’t we ditching mammograms and just doing MRIs or ultrasounds? I’ve had about enough of doctors. \\
\cmidrule(l){2-3}
& Causation, risk factors \& prevention claims
& **Strange question** A doctor once mentioned they notice a common portrait of breast cancer patients: compassionate, anxious, and taking things too personally. I went through a heartbreaking romance and severe depression right before my diagnosis, and I truly feel that emotional toll played a role.\\
\cmidrule(l){2-3}
& Safety \& effectiveness of conventional treatments
& **Is it worth taking tamoxifen for something that’s NOT breast cancer?** My oncologist prescribed me Tamoxifen, but I don’t need to take it, and it’s just going to cause more issues. \\

\bottomrule
\end{tabular}%
}

\caption{Example Reddit posts across different annotation dimensions and labels. Examples are paraphrased and anonymized for ethical reporting purposes.}
\label{tab:task_examples}
\end{table*}

\end{document}